\documentclass{article}
\usepackage{waspaa17,amsmath,amssymb,graphicx,url,times}
\usepackage{color}

\title{Diagonal RNNs in Symbolic Music Modeling}

\name{Y. Cem Subakan$^\flat$, Paris Smaragdis$^{\flat,\sharp,\natural}$  \thanks{This work was supported by NSF grant \#1453104.}}
\address{University of Illinois at Urbana-Champaign, \\
          $^\flat$Computer Science Department, $^\sharp$Electrical and Computer Engineering Department,\\
	  $^\natural$Adobe Systems, Inc. \\ 
	  \{subakan2, paris\}@illinois.edu }
\begin{document}

\maketitle

\begin{sloppy}

\begin{abstract}
In this paper, we propose a new Recurrent Neural Network (RNN) architecture. The novelty is simple: We use diagonal recurrent matrices instead of full. This results in better test likelihood and faster convergence compared to regular full RNNs in most of our experiments. We show the benefits of using diagonal recurrent matrices with popularly used LSTM and GRU architectures as well as with the vanilla RNN architecture, on four standard symbolic music datasets.    
\end{abstract}


\section{Introduction}
\label{sec:intro}
During the recent resurgence of neural networks in 2010s, Recurrent Neural Networks (RNNs) have been utilized in a variety of sequence learning applications with great success. Examples include language modeling \cite{mikolov2010}, machine translation \cite{sutskever2014}, handwriting recognition \cite{graves2013}, speech recognition \cite{hinton2013}, and symbolic music modeling \cite{Boulanger2012}. 

In this paper, we empirically show that in symbolic music modeling, using a diagonal recurrent matrix in RNNs results in significant improvement in terms of convergence speed and test likelihood.  

The inspiration for this idea comes from multivariate Gaussian Mixture Models: In Gaussian mixture modeling (or Gaussian models in general) it is known that using a diagonal covariance matrix often results in better generalization performance, increased numerical stability and reduced computational complexity \cite{Reynolds2015, Alpaydin2010}. We adapt this idea to RNNs by using diagonal recurrent matrices.

We investigate the consequences of using diagonal recurrent matrices for the vanilla RNNs, and for more popular Long Short Term Memory Networks (LSTMs) \cite{Hochreiter1997, Graves2005} and Gated Recurrent Units (GRUs) \cite{Chung2014}. We empirically observe that using diagonal recurrent matrices results in an improvement in convergence speed in training and the resulting test likelihood for all three models, on four standard symbolic music modeling datasets.

\section{Recurrent Neural Networks}

\label{sec:rnn} 
The vanilla RNN (VRNN) recursion is defined as follows: 
\begin{align}
	h_t =& \sigma_1 ( W h_{t-1} + U x_t + b ),  \label{eq:vrnn} 
\end{align}
where $h_t \in \mathbb R^K$ is the hidden state vector with $K$ hidden units, and $x_t\in \mathbb R^L$ is the input vector at time $t$ (which has length $L$). The $U \in \mathbb R^{K \times L}$ is the input matrix that transforms the input $x_t$ from an $L$ to $K$ dimensional space and $W \in \mathbb R^{K \times K}$ is the recurrent matrix (factor) that transforms the previous state. Finally, $b\in \mathbb R^K$ is the bias vector. Note that, in practice this recursion is either followed by an output stage on top of $h_t$ to get the outputs as $y_t = \sigma_2(V h_t)  \in \mathbb R^L$, or another recursion to obtain a multi-layer recurrent neural network. The hidden layer non-linearity $\sigma_1(.)$ is usually chosen as hyperbolic tangent. The choice of the output non-linearity $\sigma_2(.)$ is dependent on the application, and is typically softmax or sigmoid function.  

Despite its simplicity, RNN in its original form above is usually not preferred in practice due to the well known gradient vanishing problem \cite{Pascanu2013}. People often use the more involved architectures such as LSTMs and GRUs, which alleviate the vanishing gradient issue using gates which filter the information flow to enable the modeling of long-term dependencies. 

\subsection{LSTM and GRU}
The GRU Network is defined as follows:  
\begin{align}
	f_t =& \sigma(W_f h_{t-1} + U_f x_t ), \notag \\
	w_t =& \sigma(W_w  h_{t-1} + U_w x_t ), \notag \\  
	c_t =& \tanh(W (h_{t-1}\odot w_t) + U x_t), \notag \\ 
	h_t =& h_{t-1} \odot f_t + (1-f_t) \odot c_t,  \label{eq:gru}  
\end{align}
where $\odot$ denotes element-wise (Hadamard) product, $\sigma(.)$ is the sigmoid function, $f_t \in \mathbb R^{K}$ is the forget gate, and $w_t \in \mathbb R^{K}$ is the write gate: If $f_t$ is a zeros vector, the current state $h_t$ depends solely on the candidate vector $c_t$. On the other extreme where $f_t$ is a ones vector, the state $h_{t-1}$ is carried over unchanged to $h_t$. Similarly, $w_t$ determines how much $h_{t-1}$ contributes to the candidate state $c_t$. Notice that if $w_t$ is a ones vector and $f_t$ is a zeros vector, the GRU architecture reduces to the VRNN architecture in Equation \eqref{eq:vrnn}. Finally, note that we have omitted the biases in the equations for $f_t$, $w_t$, and $c_t$ to reduce the notation clutter. We will omit the bias terms also in the rest of this paper.\\ 

The LSTM Network is very much related to the GRU network above. In addition to the gates in GRU, there is the output gate $o_t$ to control the output of the RNN, and the forget gate is decoupled into gates $f_t$ and $w_t$, which blend the previous state and the candidate state $c_t$: 
\begin{align}
	f_t =& \sigma(W_f h_{t-1} + U_f x_t ), \notag \\
	w_t =& \sigma(W_w  h_{t-1} + U_w x_t ), \notag \\  
	o_t =& \sigma(W_o  h_{t-1} + U_o x_t ), \notag \\  
	c_t =& \tanh(W h_{t-1} + U x_t), \notag \\ 
	h'_t =& h_{t-1}' \odot f_t + w_t \odot c_t, \notag \\
	h_t =& o_t \odot \tanh(h'_t), \label{eq:gru}  
\end{align}
Also notice the application of the tangent hyperbolic on $h'_t$ before yielding the output. This prevents the output from assuming values with too large magnitudes. In \cite{Greff2015} it is experimentally shown that this output non-linearity is crucial for the LSTM performance. 

\subsection{Diagonal RNNs}
  We define the Diagonal RNN as an RNN with diagonal recurrent matrices. The simplest case is obtained via the modification of the VRNN. After the modification, the VRNN recursion becomes the following:
\begin{align}
	h_t =& \sigma_1 ( W \odot h_{t-1} + U x_t ),  \label{eq:dvrnn} 
\end{align}
where this time the recurrent term $W$ is a length $K$ vector, instead of a $K\times K$ matrix. Note that element wise multiplying the previous state $h_{t-1}$ with the $W$ vector is equivalent to having a matrix-vector multiplication $W_{diag}h_{t-1}$ where $W_{diag}$ is a diagonal matrix, with diagonal entries set to the $W$ vector, and hence the name for Diagonal RNNs. For the more involved GRU and LSTM architectures, we also modify the recurrent matrices of the gates. This results in the following network architecture for GRU: 
\begin{align}
	f_t =& \sigma(W_f \odot h_{t-1} + U_f x_t ), \notag \\
	w_t =& \sigma(W_w \odot  h_{t-1} + U_w x_t ), \notag \\  
	c_t =& \tanh(W \odot h_{t-1}\odot w_t + U x_t), \notag \\ 
	h_t =& h_{t-1} \odot f_t + (1-f_t) \odot c_t,  \label{eq:dgru}  
\end{align}
where $W_f,W_w,W \in \mathbb R^K$. Similarly for LSTM, we obtain the following: 
\begin{align}
	f_t =& \sigma(W_f \odot h_{t-1} + U_f x_t ), \notag \\
	w_t =& \sigma(W_w \odot h_{t-1} + U_w x_t ), \notag \\  
	o_t =& \sigma(W_o \odot  h_{t-1} + U_o x_t ), \notag \\  
	c_t =& \tanh(W \odot h_{t-1} + U x_t), \notag \\ 
	h'_t =& h_{t-1}' \odot f_t + w_t \odot c_t, \notag \\
	h_t =& o_t \odot \tanh(h'_t), \label{eq:dlstm}  
\end{align}
where again $W_f, W_w, W_o, W \in \mathbb R^{K}$. One more thing to note is that the total number of trainable parameters in this model scales as $\mathcal{O}(K)$ and not $\mathcal{O}(K^2)$ like the regular full architectures, which implies lower memory and computation requirements.   

\subsection{Intuition on Diagonal RNNs}
In order to gain some insight on how diagonal RNNs differ from regular full RNNs functionally, let us unroll the VRNN recursion in Equation \ref{eq:vrnn}:
\begin{align}
	& h_t = \sigma ( W \sigma ( W h_{t-2} + U x_{t-1} ) + U x_t ) \label{eq:vrnn-rec} \\ 
	    =& \sigma ( W \sigma ( W \sigma ( W h_{t-3} + U x_{t-2} ) + Ux_{t-1} ) + U x_t ) \notag \\
	    =& \sigma ( W \sigma ( W \sigma( \dots W\sigma( W h_0 + U x_1) +  \dots ) + Ux_{t-1} ) + U x_t ) \notag 
\end{align}
So, we see that the RNN recursion forms a mapping from $x_{1:t} = (x_1,\dots,x_{t-1},x_t)$ to $h_t$. That is, the state $h_t$ is a function of all past inputs and the current input. To get an intuition on how the recurrent matrix $W$ interacts with the inputs $x_{1:t}$ functionally, we can temporarily ignore the $\sigma(.)$ non-linearities:   
\begin{align} 
	h_t =& W^t h_0 + W^{t-1} U x_1 + W^{t-2} U x_2 + \dots +  U x_t \notag \\
	=& W^t h_0 + \sum_{k = 1}^t W^{t-k} U x_k. \label{eq:vrnn-linrec} 
\end{align}
Although this equation sacrifices from generality, it gives a notion on how the $W$ matrix effects the overall transformation: After the input transformation via the $U$ matrix, the inputs are further transformed via multiple application of $W$ matrices: The exponentiated $W$ matrices act as ``weights'' on the inputs. Now, the question is, why are the weights applied via $W$ are the way they are? The input transformations via $U$ are sensible since we want to project our inputs to a $K$ dimensional space. But the transformations via recurrent weights $W$ are rather arbitrary as there are multiple plausible forms for $W$.

We can now see that a straightforward alternative to the RNN recursion in equation \eqref{eq:vrnn} is considering linear transformations via diagonal, scalar and constant alternatives for the recurrent matrix $W$, similar to the different cases for Gaussian covariance matrices \cite{Alpaydin2010}. In this paper, we explore the diagonal alternative to the full $W$ matrices.


One last thing to note is that using a diagonal matrix does not completely eliminate the ability of the neural network to model inter-dimensional correlations since the projection matrix $U$gets applied on each input $x_t$, and furthermore, most networks typically has a dense output layer.   

\section{Experiments}
We trained VRNNs, LSTMs and GRUs with full and diagonal recurrent matrices on the symbolic midi music datasets. We downloaded the datasets from \url{http://www-etud.iro.umontreal.ca/~boulanni/icml2012} which are originally used in the paper \cite{Boulanger2012}. The learning goal is to predict the next frame in a given sequence using the past frames. All datasets are divided into training, test, and validation sets. The performance is measured by the per-frame negative log-likelihood on the sequences in the test set.  

The datasets are ordered in increasing size as, JSB Chorales, Piano-Midi, Nottingham and MuseData. We did not apply any transposition to center the datasets around a key center, as this is an optional preprocessing as indicated in \cite{Boulanger2012}. We used the provided piano roll sequences provided in the aforementioned url, and converted them into binary masks where the entry is one if there is a note played in the corresponding pitch and time. We also eliminated the pitch bins for which there is no activity in a given dataset. Due to the large size of our experiments, we limited the maximum sequence length to be 200 (we split the sequences longer than 200 into sequences of length 200 at maximum) to take advantage of GPU parallelization, as we have noticed that this operation does not alter the results significantly. 

We randomly sampled 60 hyper-parameter configurations for each model in each dataset, and for each optimizer. We report the test accuracies for the top 6 configurations, ranked according to their performance on the validation set. For each random hyper-parameter configuration, we trained the given model for 300 iterations. We did these experiments for two different optimizers. Overall, we have 6 different models (VRNN full, VRNN diagonal, LSTM full, LSTM diagonal, GRU full, GRU diagonal), and 4 different datasets, and 2 different optimizers, so this means that we obtained $6\times4\times2\times 60 = 2880$ training runs, 300 iterations each. We trained our models on Nvidia Tesla K80 GPUs. 

As optimizers, we used the Adam optimizer \cite{Kingma2014} with the default parameters as specified in the corresponding paper, and RMSprop \cite{RMSProp}. We used a sigmoid output layer for all models. We used mild dropout in accordance with \cite{Zaremba2014} with keep probability $0.9$ on the input and output of all layers. We used Xavier initialization \cite{Glorot2010} for all cases. The sampled hyper-parameters and corresponding ranges are as follows: 

\begin{itemize}
	\item Number of hidden layers: Uniform Samples from \{2,3\}.
	\item Number of hidden units per hidden layer: Uniform Samples from \{50,\dots,300\} for LSTM, and uniform samples from \{50,\dots,350\} for GRU, and uniform samples from \{50,\dots,400\} for VRNN. 
	\item Learning rate: Log-uniform samples from the range $[10^{-4}, 10^{-2}]$. 
	\item Momentum (For RMS-Prop): Uniform samples from the range $[0,1]$.
\end{itemize}

As noted in the aforementioned url, we used the per-frame negative log-likelihood measure to evaluate our models. The negative log-likelihood is essentially the cross-entropy between our predictions and the ground truth. Per frame negative log-likelihood is given by the following expression: 
\begin{align}
  \text{Per Frame Negative Log-Likelihood} = -\frac{1}{T}\sum_{t=1}^T y_t \log \hat y_t, \notag
\end{align}
where $y_t$ is the ground truth for the predicted frames and $\hat y_t$ is the output of our neural network, and $T$ is the number of time steps (frames) in a given sequence. 

\newcommand{\fullcolor}{cyan }
\newcommand{\diagcolor}{black }
\newcommand{\Fullcolor}{Cyan }
\newcommand{\Diagcolor}{Black }

In Figures \ref{fig:JSB}, \ref{fig:Piano-midi}, \ref{fig:Nottingham}, and \ref{fig:MuseData} we show the training iterations vs negative test log-likelihoods for top 6 hyperparameter configurations on JSB Chorales, Piano-midi, Nottingham and MuseData datasets, respectively. That is, we show the negative log-likelihoods obtained on the test set with respect to the training iterations, for top 6 hyper-parameter configurations ranked on the validation set according to the performance attained at the last iteration. The top rows show the training iterations for the Adam optimizer and the bottom rows show the iterations for the RMSprop optimizer. The curves show the negative log-likelihood averaged over the top 6 configurations, where \fullcolor curves are for full model and \diagcolor curves are for diagonal models. We use violin plots, which show the distribution of the test negative log-likelihoods of the top 6 configurations. We also show the average number of parameters used in the models corresponding to top 6 configurations in the legends of the figures. The minimum negative log-likelihood values obtained with each model using Adam and RMSprop optimizers are summarized in Table \ref{table:minlogls}.


%

\begin{table*}
  \caption{Minimum Negative Log-Likelihoods on Test Data (Lower is better) with Adam and RMSProp optimizers. \textbf F stands for Full models and \textbf D stands for Diagonal models. }
  \label{table:minlogls}
  \centering
  \begin{tabular}{|l|c|c|c|c|c|c|} 
    \hline
    \textbf{Dataset/Optimizer} & \textbf{RNN-F} & \textbf{RNN-D} &\textbf{LSTM-F} & \textbf{LSTM-D} & \textbf{GRU-F} & \textbf{GRU-D}\\ \hline \hline 
	JSB Chorales/Adam & 8.91 & \textbf{8.12} & 8.56& 8.23& 8.64& {8.21}\\ \hline
	Piano-Midi/Adam & 7.74& \textbf{7.53} & 8.83 & 7.59 & 8.28 & 7.54 \\ \hline
	Nottingham/Adam & \textbf{3.57} & 3.69 &3.90 & 3.74& \textbf{3.57} & 3.61 \\ \hline 
	MuseData/Adam & 7.82 & 7.26 &  8.96 & \textbf{7.08}& 7.52& 7.20 \\ \hline
	JSB Chorales/RMSprop & 8.72 & 8.22 &8.51 &\textbf{8.14} &8.53 & 8.22 \\ \hline
	Piano-Midi/RMSprop & 7.65 & 7.51 & 7.84 & 7.49  &7.62 & \textbf{7.48}  \\ \hline
	Nottingham/RMSprop & \textbf{3.40} & 3.67  & 3.54 & 3.65 & 3.45  & 3.62  \\ \hline 
	MuseData/RMSprop & 7.14 & 7.23 & 7.20 &7.09 & 7.11 &\textbf{6.96} \\ \hline
  \end{tabular}
\end{table*}

We implemented all models in Tensorflow \cite{Tensorflow2015}, and our code can be downloaded from our github page \url{https://github.com/ycemsubakan/diagonal_rnns}. All of the results presented in this paper are reproducible with the provided code. 
\newcommand{\sz}{0.156}
\newcommand{\lc}{0.85}
\newcommand{\rc}{0.87}
\newcommand{\bc}{0.9}
\newcommand{\uc}{0.7}

\begin{figure}[ht]
  \centering
  \includegraphics[trim = {\lc cm \bc cm \rc cm \uc cm},clip, width = \sz\textwidth]{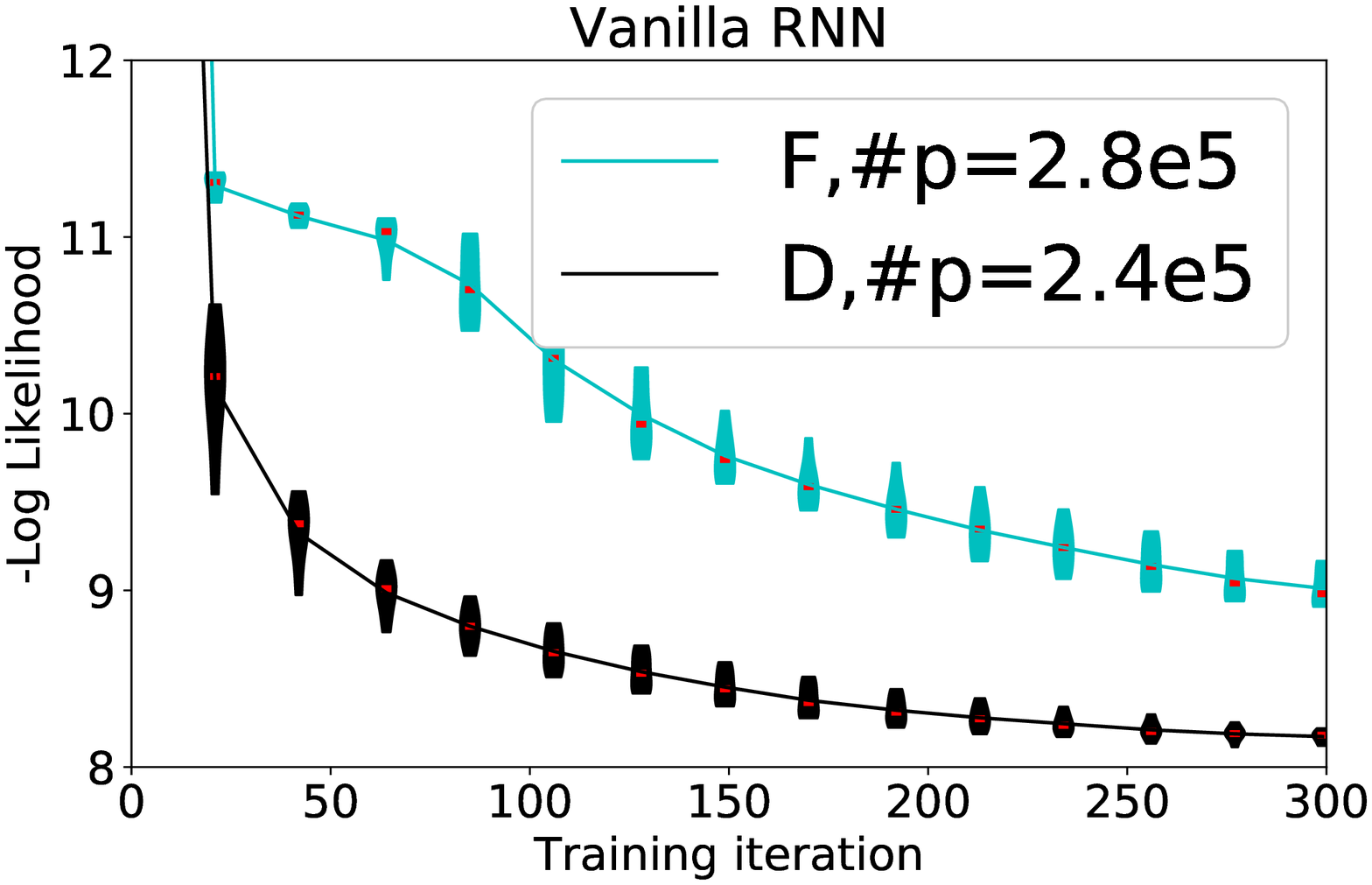}  
  \includegraphics[trim = {\lc cm \bc cm \rc cm \uc cm},clip, width = \sz\textwidth]{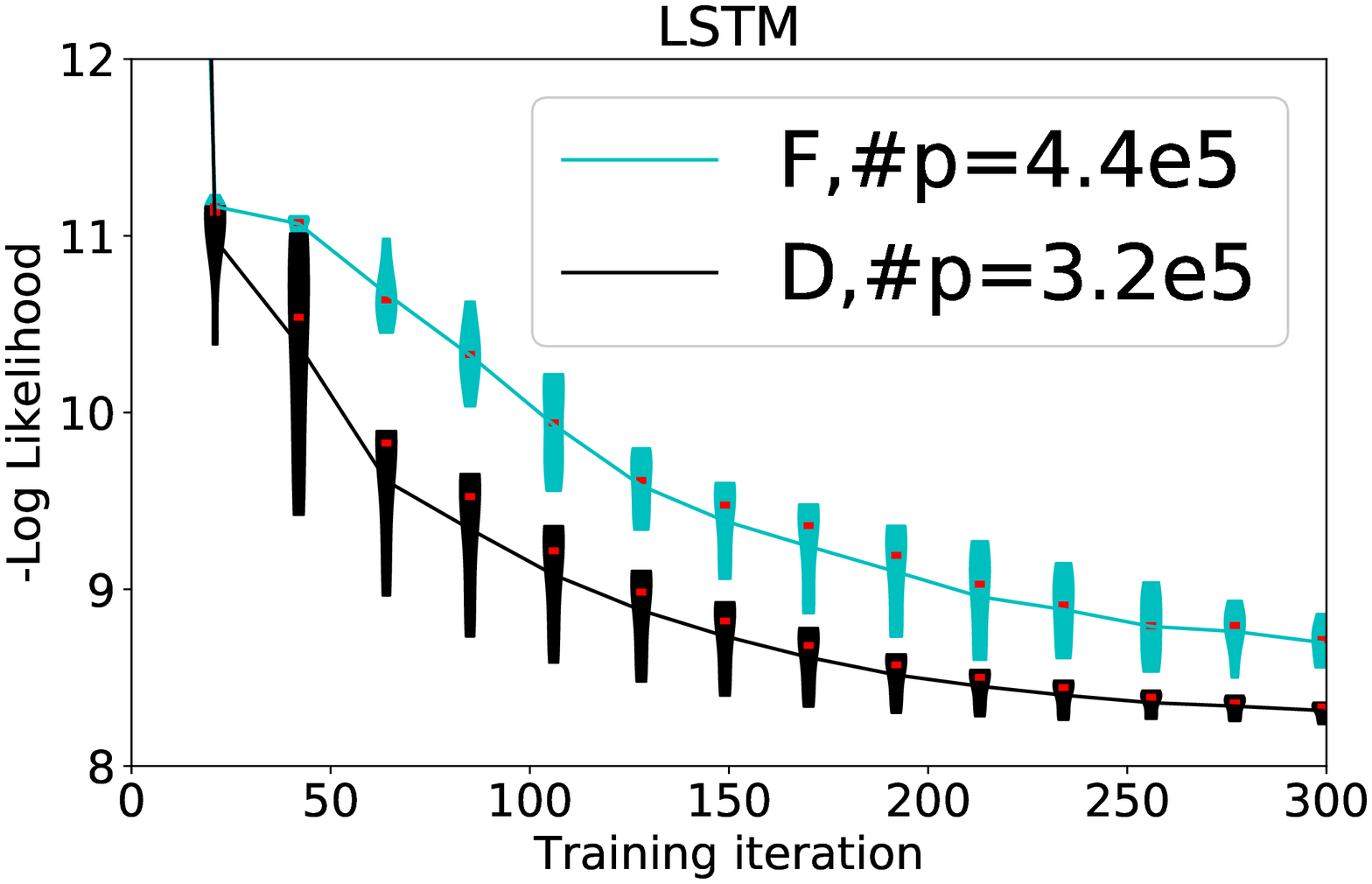} 
  \includegraphics[trim = {\lc cm \bc cm \rc cm \uc cm},clip, width = \sz\textwidth]{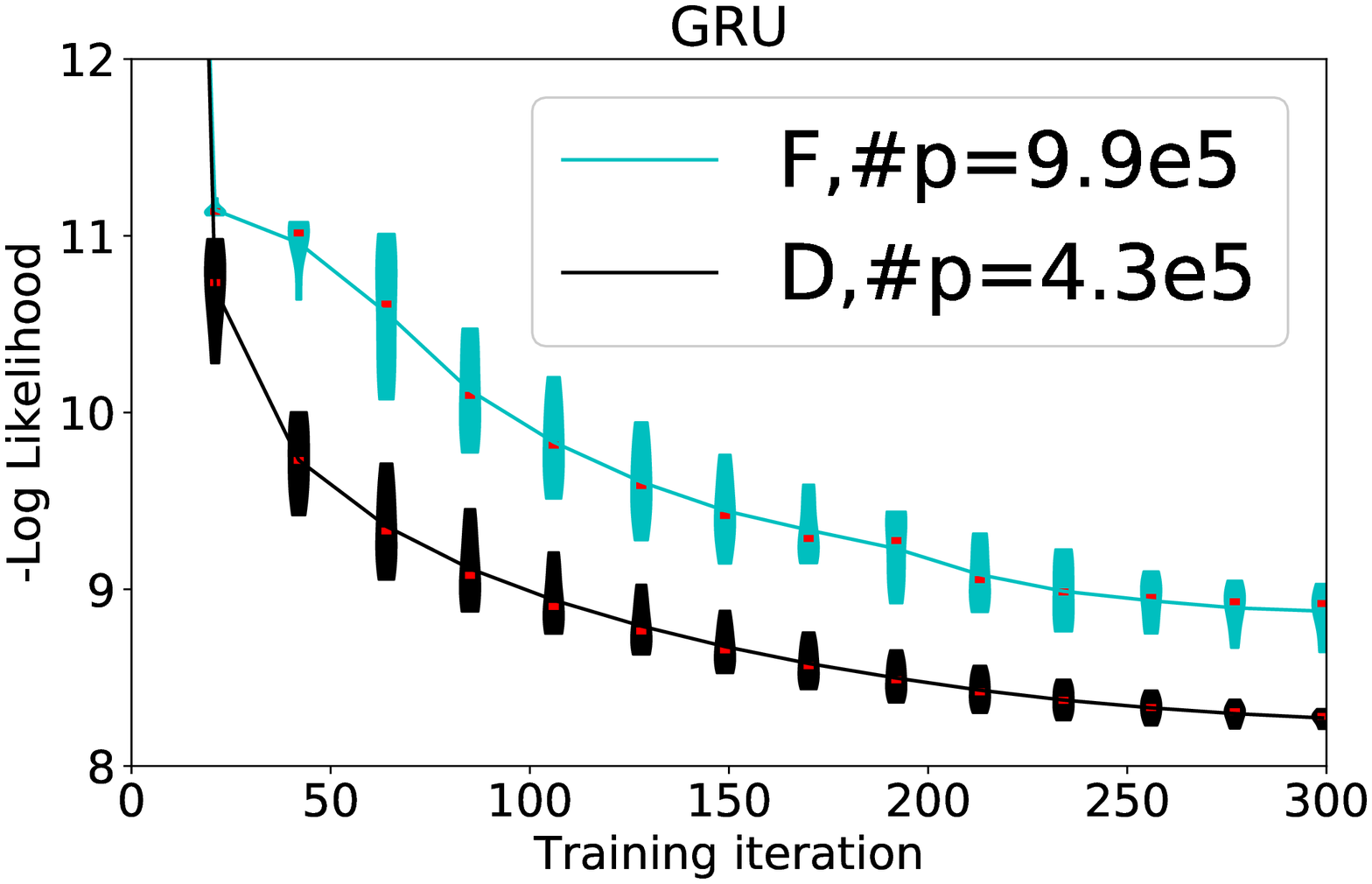} 

  \includegraphics[trim = {\lc cm \bc cm \rc cm \uc cm},clip, width = \sz\textwidth]{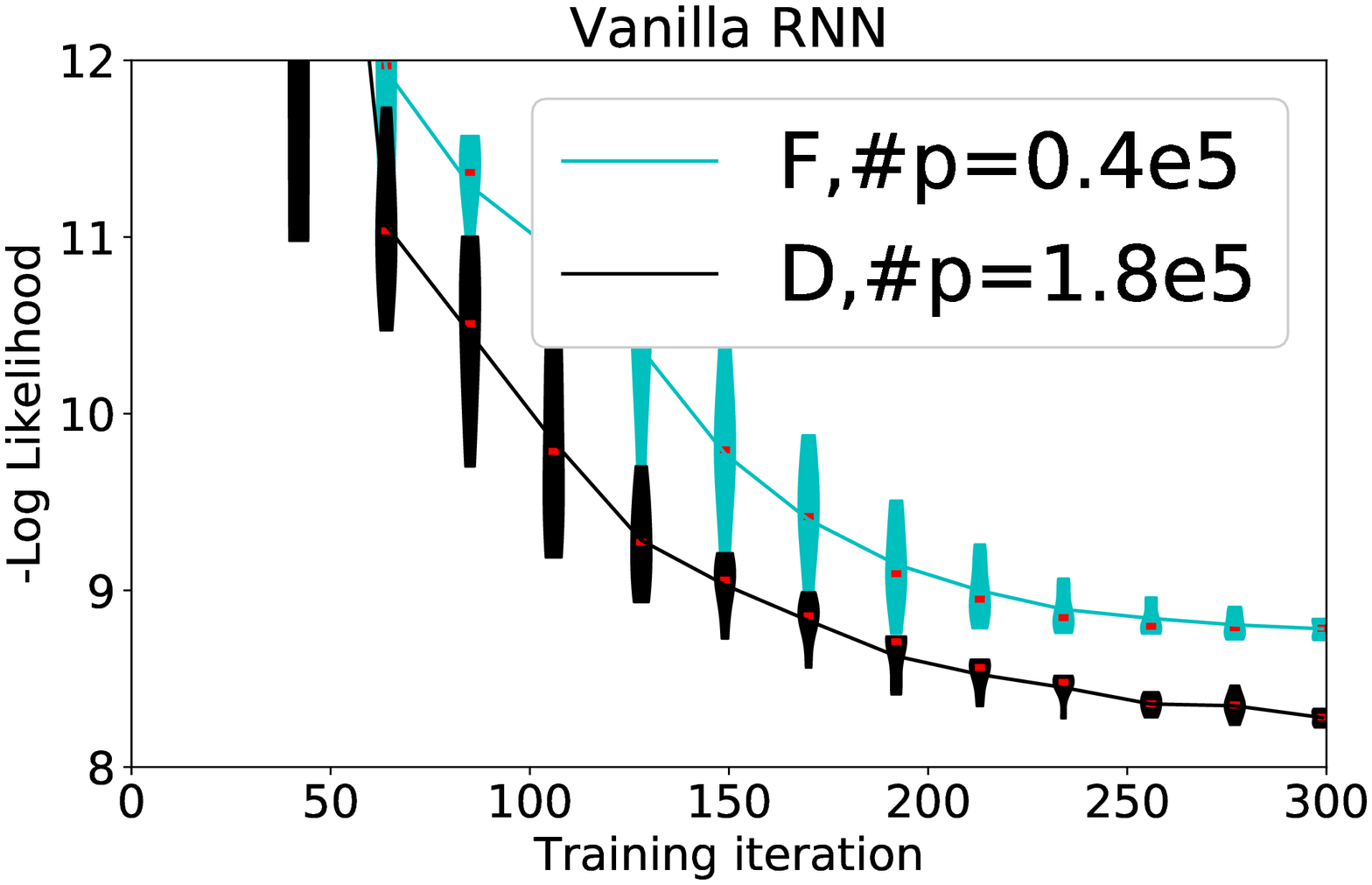}  
  \includegraphics[trim = {\lc cm \bc cm \rc cm \uc cm},clip, width = \sz\textwidth]{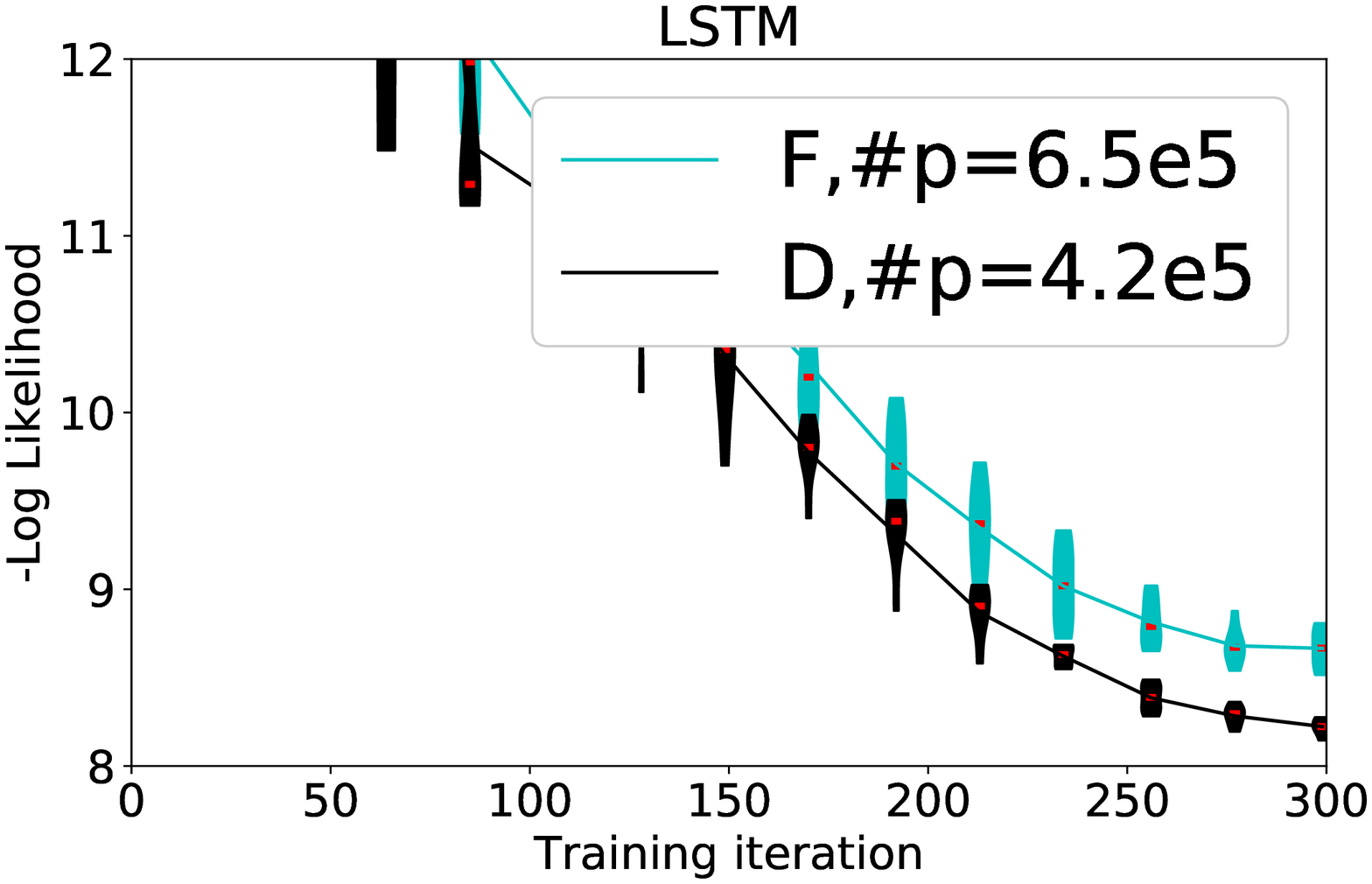} 
  \includegraphics[trim = {\lc cm \bc cm \rc cm \uc cm},clip, width = \sz\textwidth]{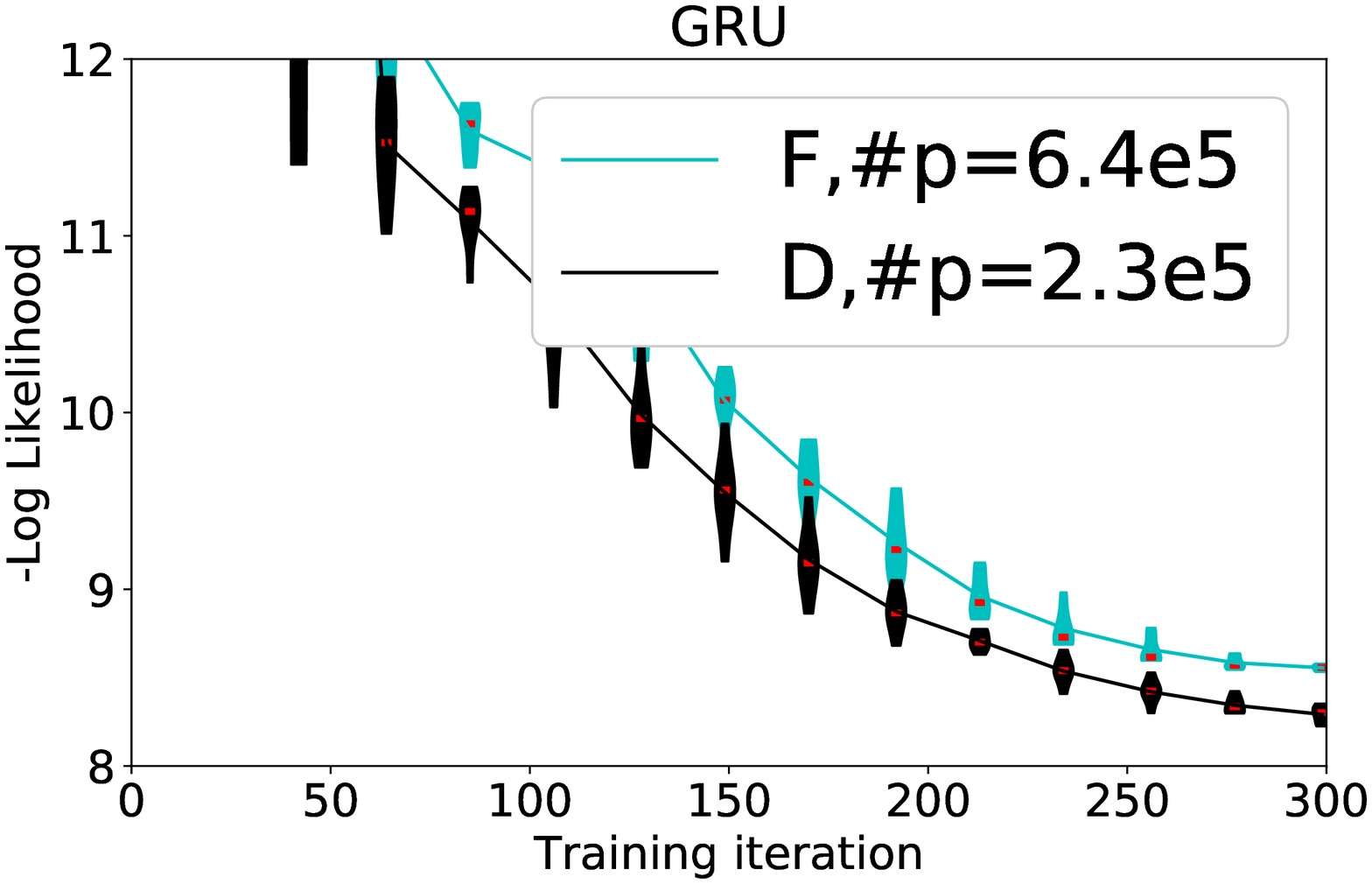} 
  \vspace{-0.2cm}
  \caption{Training iterations vs test negative log-likelihoods on JSB Chorales dataset for full and diagonal models. Top row is for the Adam optimizer and the bottom row is for RMSProp. \Diagcolor curves are for the diagonal models and \fullcolor (gray in grayscale) curves are for full (regular) models. Left column is for VRNN, middle column is for LSTM and right column is for GRU. Legends show the average number of parameters used by top 6 models (F is for Full, D is for Diagonal models). This caption also applies to Figures \ref{fig:Piano-midi}, \ref{fig:Nottingham}, \ref{fig:MuseData}, with corresponding datasets.}
  \label{fig:JSB} 
\end{figure}

\begin{figure}[ht]
  \centering
  \includegraphics[trim = {\lc cm \bc cm \rc cm \uc cm},clip, width = \sz\textwidth]{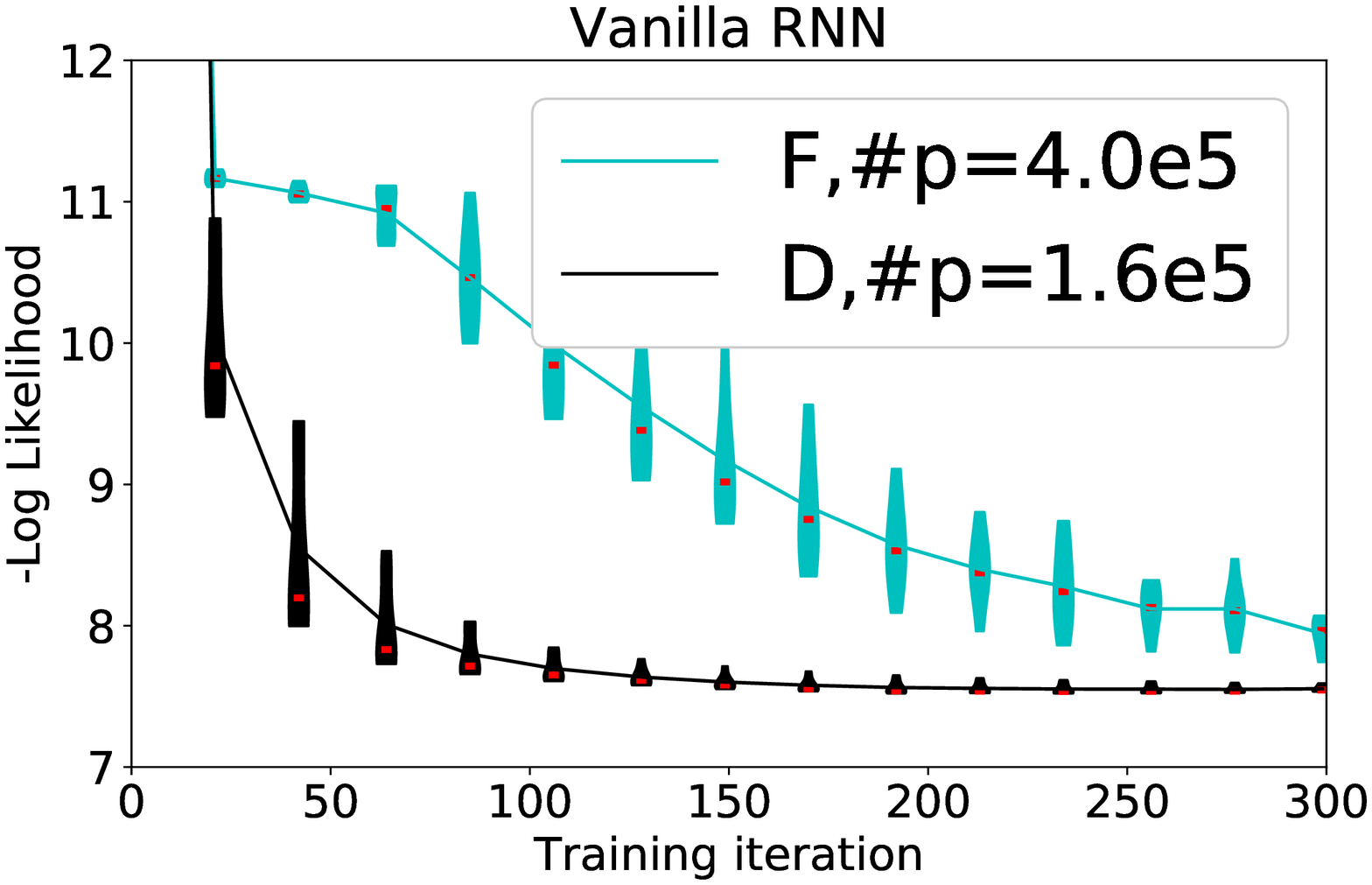}  
  \includegraphics[trim = {\lc cm \bc cm \rc cm \uc cm},clip, width = \sz\textwidth]{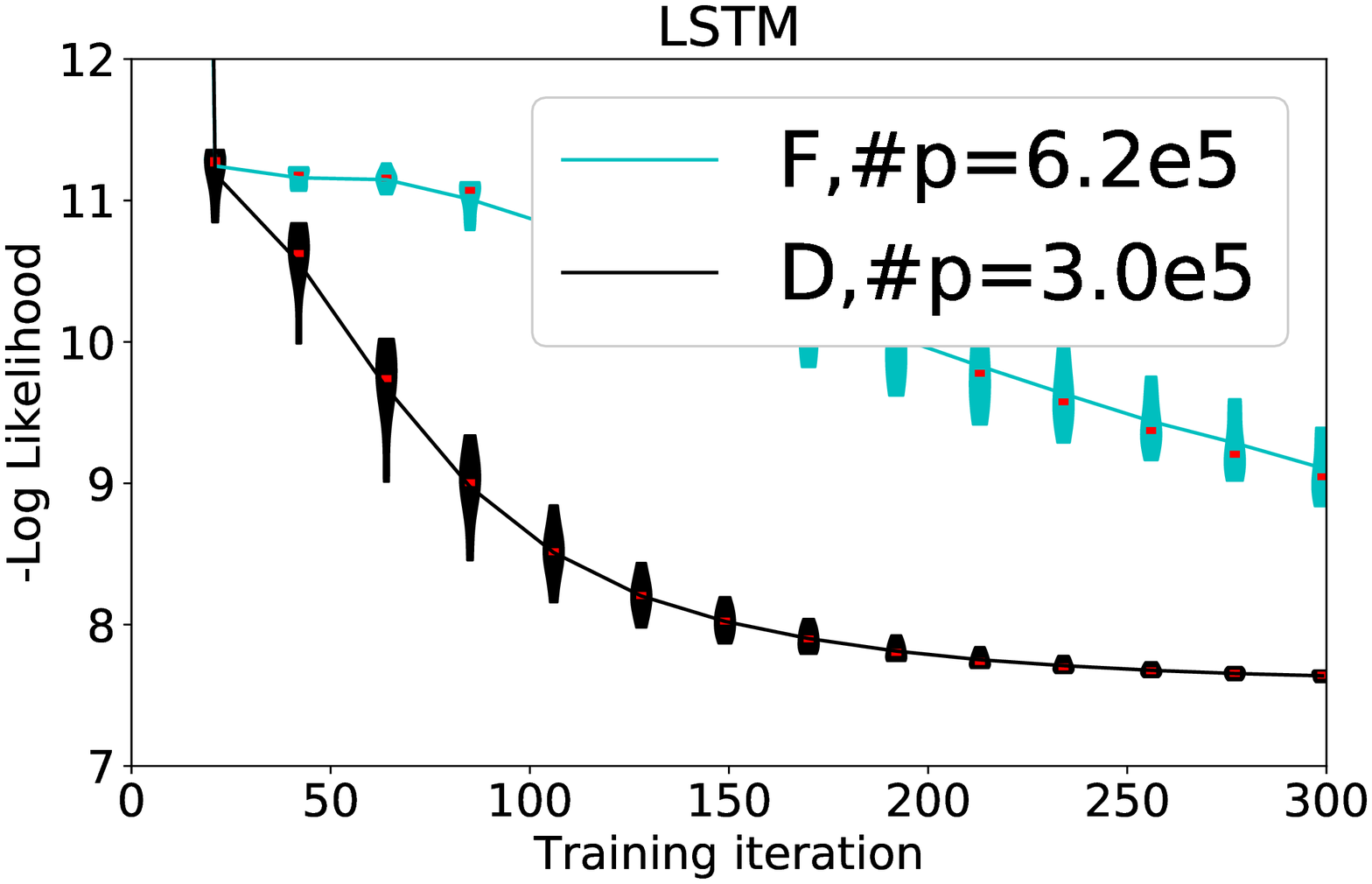} 
  \includegraphics[trim = {\lc cm \bc cm \rc cm \uc cm},clip, width = \sz\textwidth]{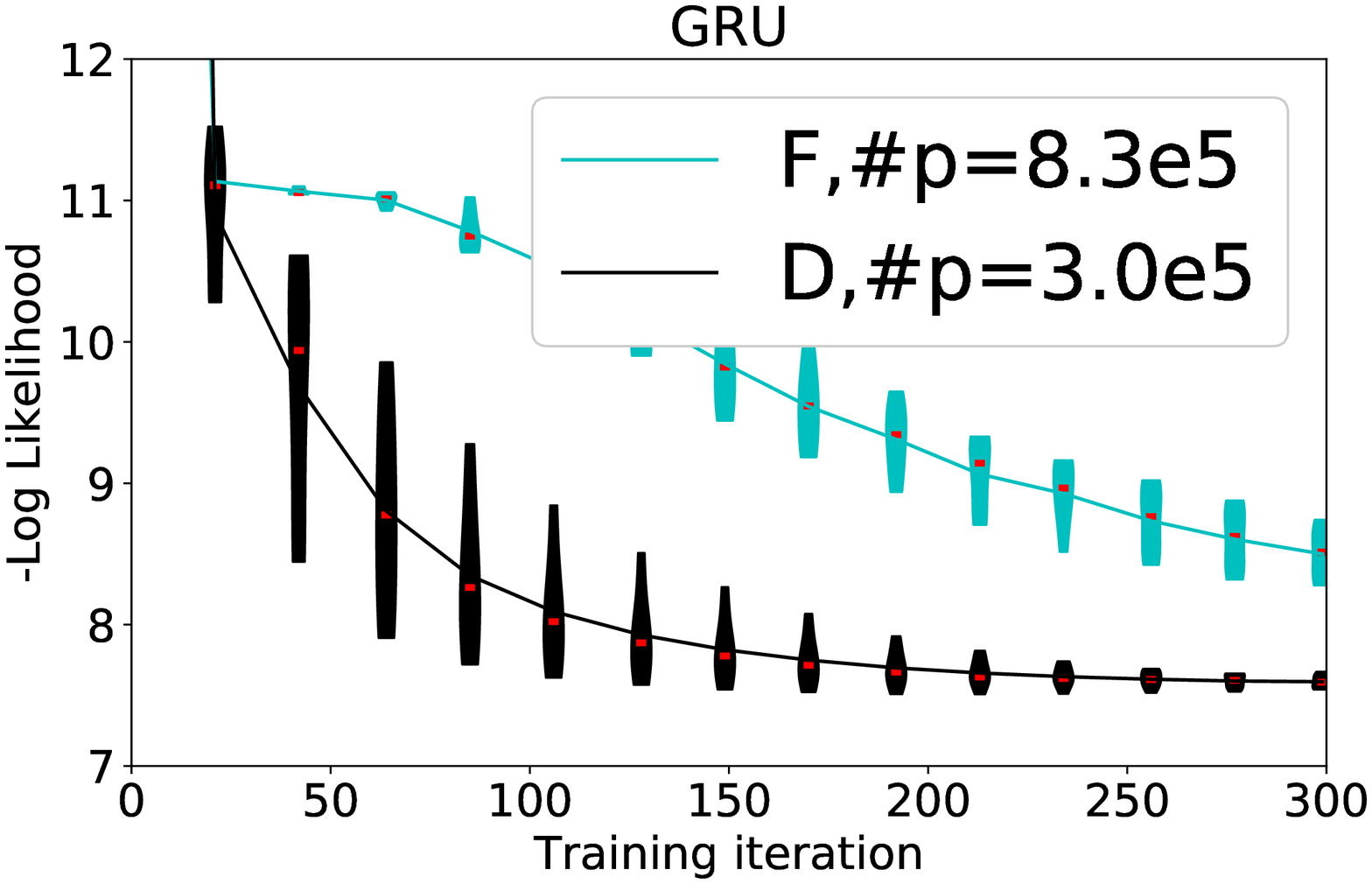} 

  \includegraphics[trim = {\lc cm \bc cm \rc cm \uc cm},clip, width = \sz\textwidth]{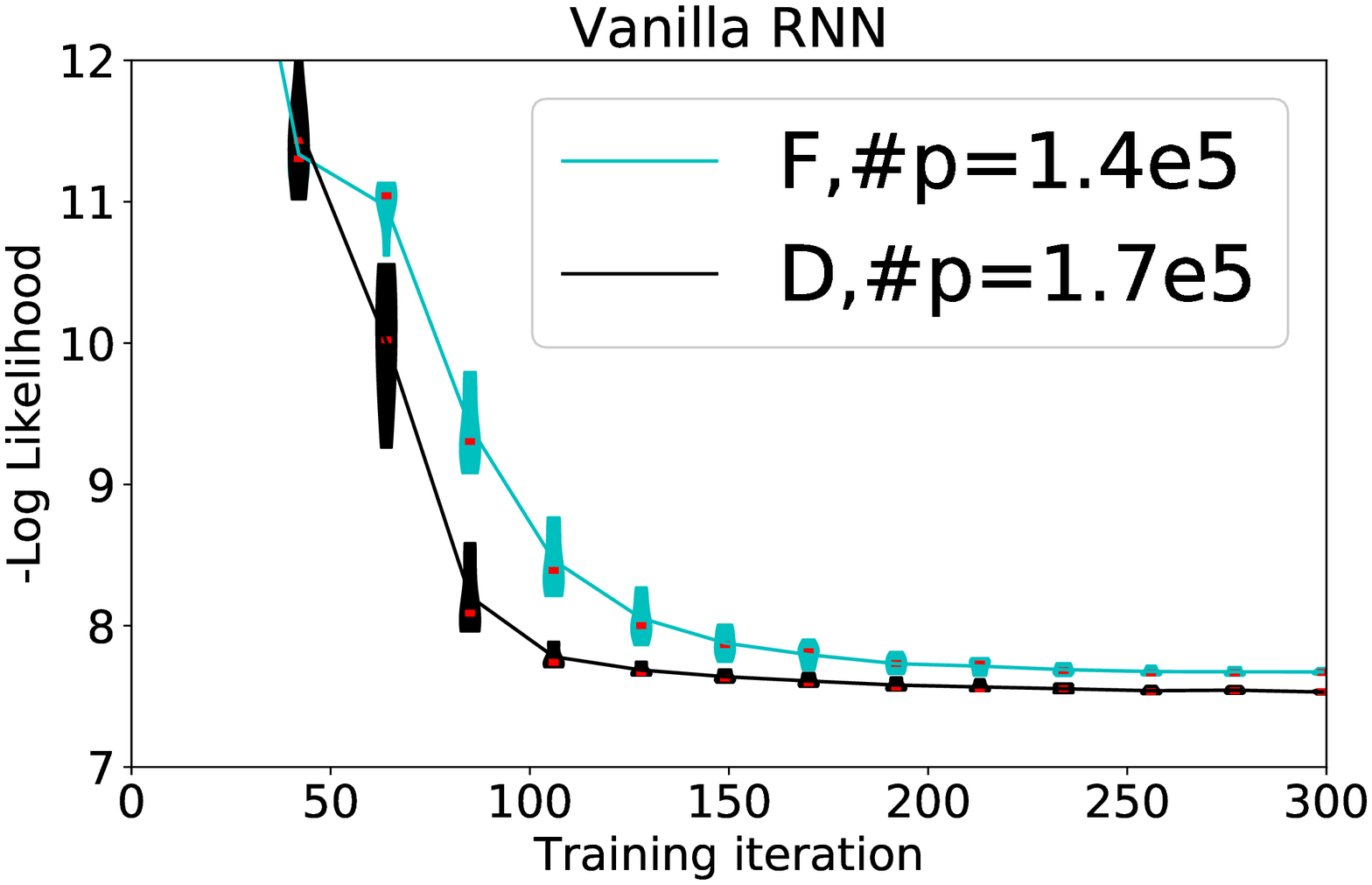}  
  \includegraphics[trim = {\lc cm \bc cm \rc cm \uc cm},clip, width = \sz\textwidth]{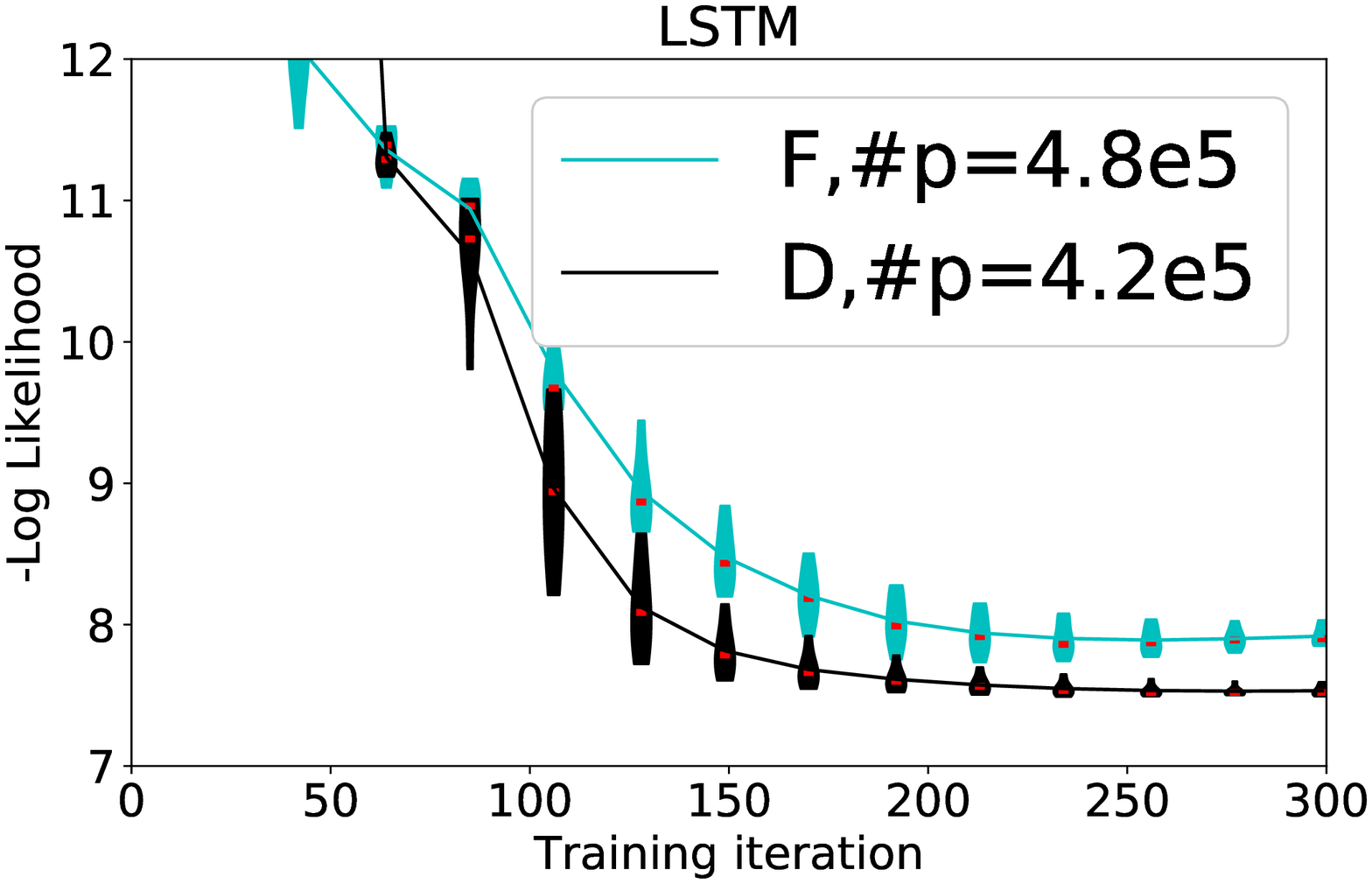} 
  \includegraphics[trim = {\lc cm \bc cm \rc cm \uc cm},clip, width = \sz\textwidth]{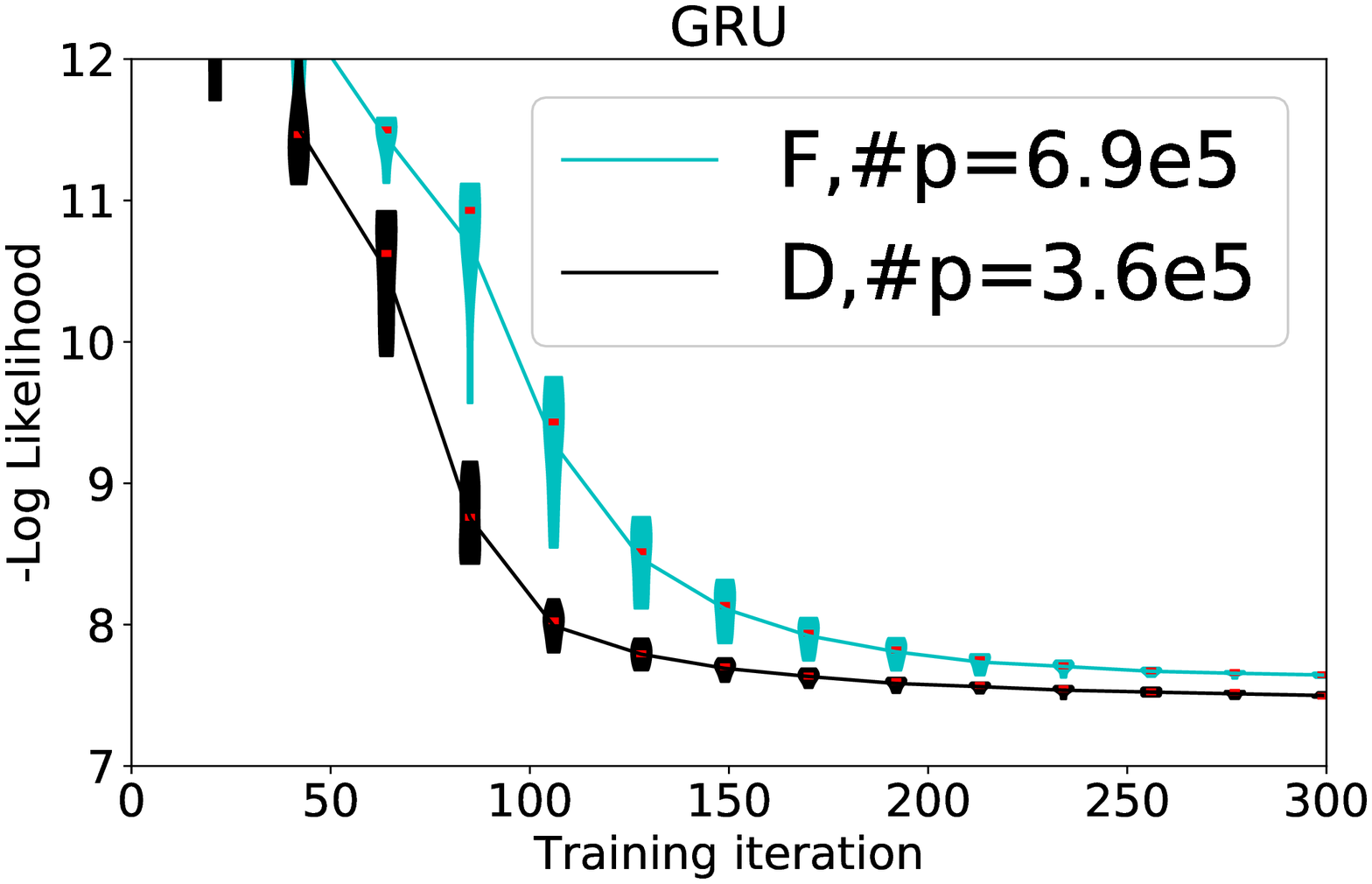} 
 
  \vspace{-0.2cm}
  \caption{Training iterations vs test negative log-likelihoods on Piano-midi dataset.}
  \label{fig:Piano-midi}
\end{figure}

\begin{figure}[ht]
  \centering
  \includegraphics[trim = {\lc cm \bc cm \rc cm \uc cm},clip, width = \sz\textwidth]{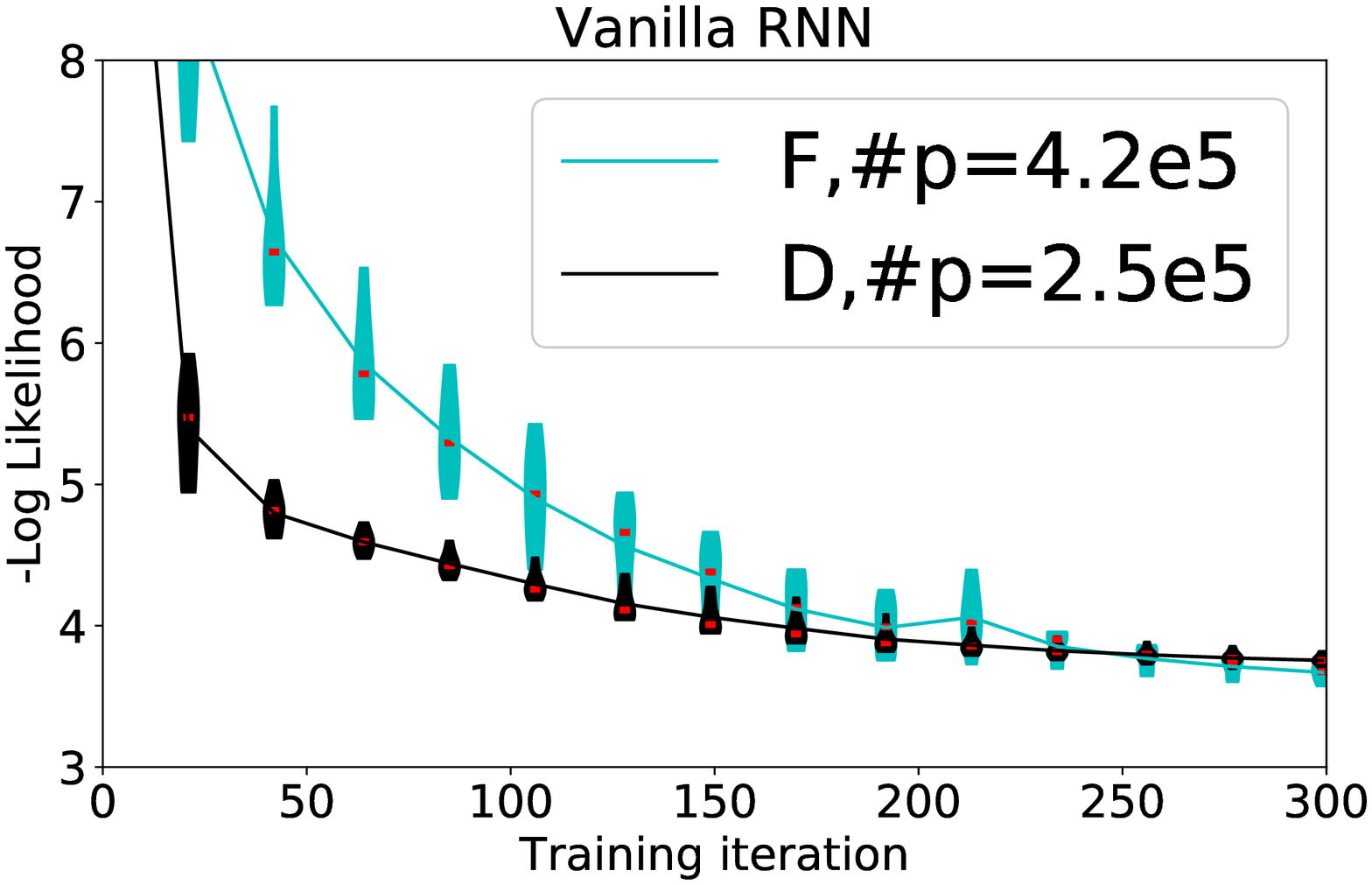}  
  \includegraphics[trim = {\lc cm \bc cm \rc cm \uc cm},clip, width = \sz\textwidth]{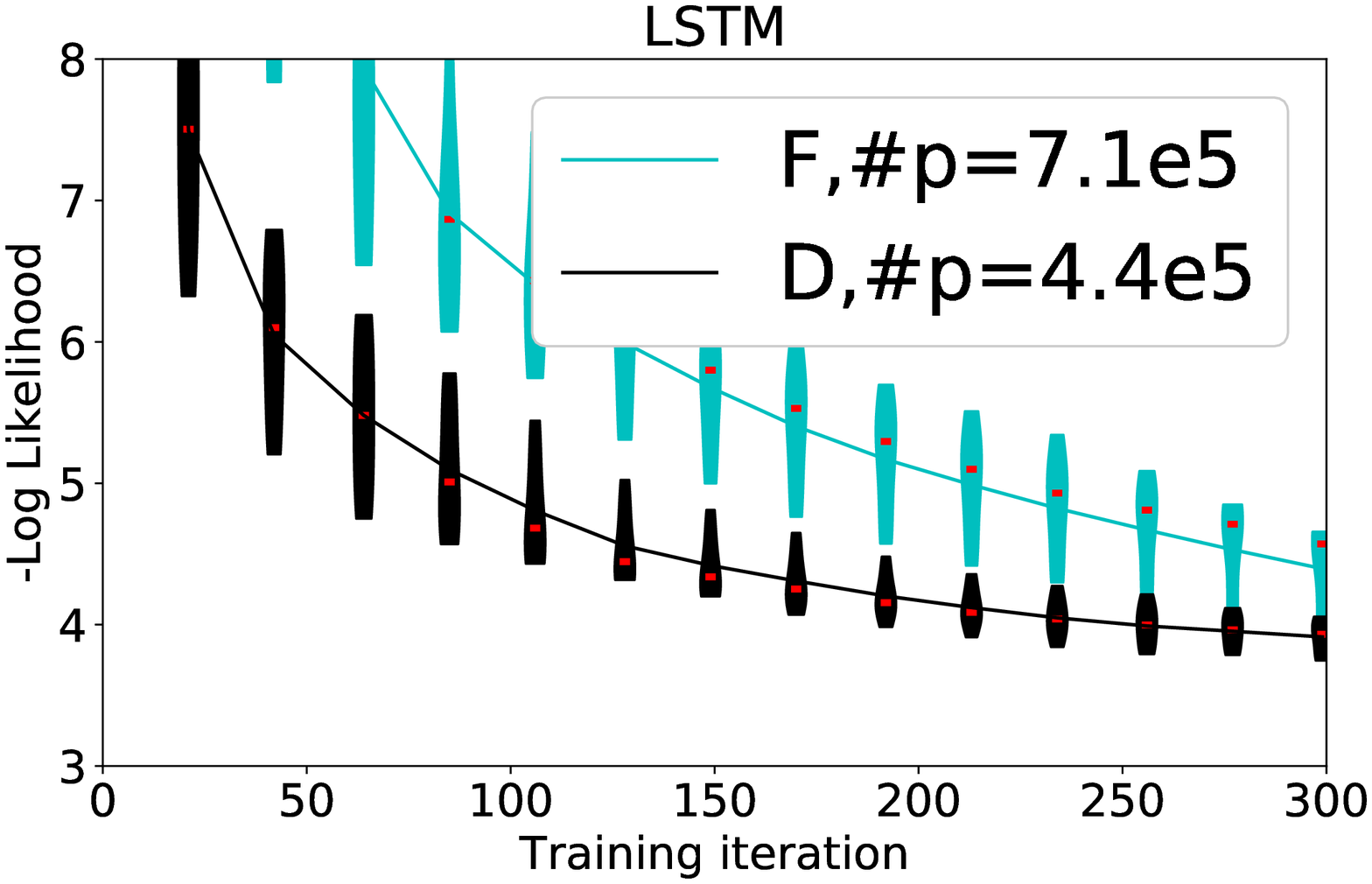} 
  \includegraphics[trim = {\lc cm \bc cm \rc cm \uc cm},clip, width = \sz\textwidth]{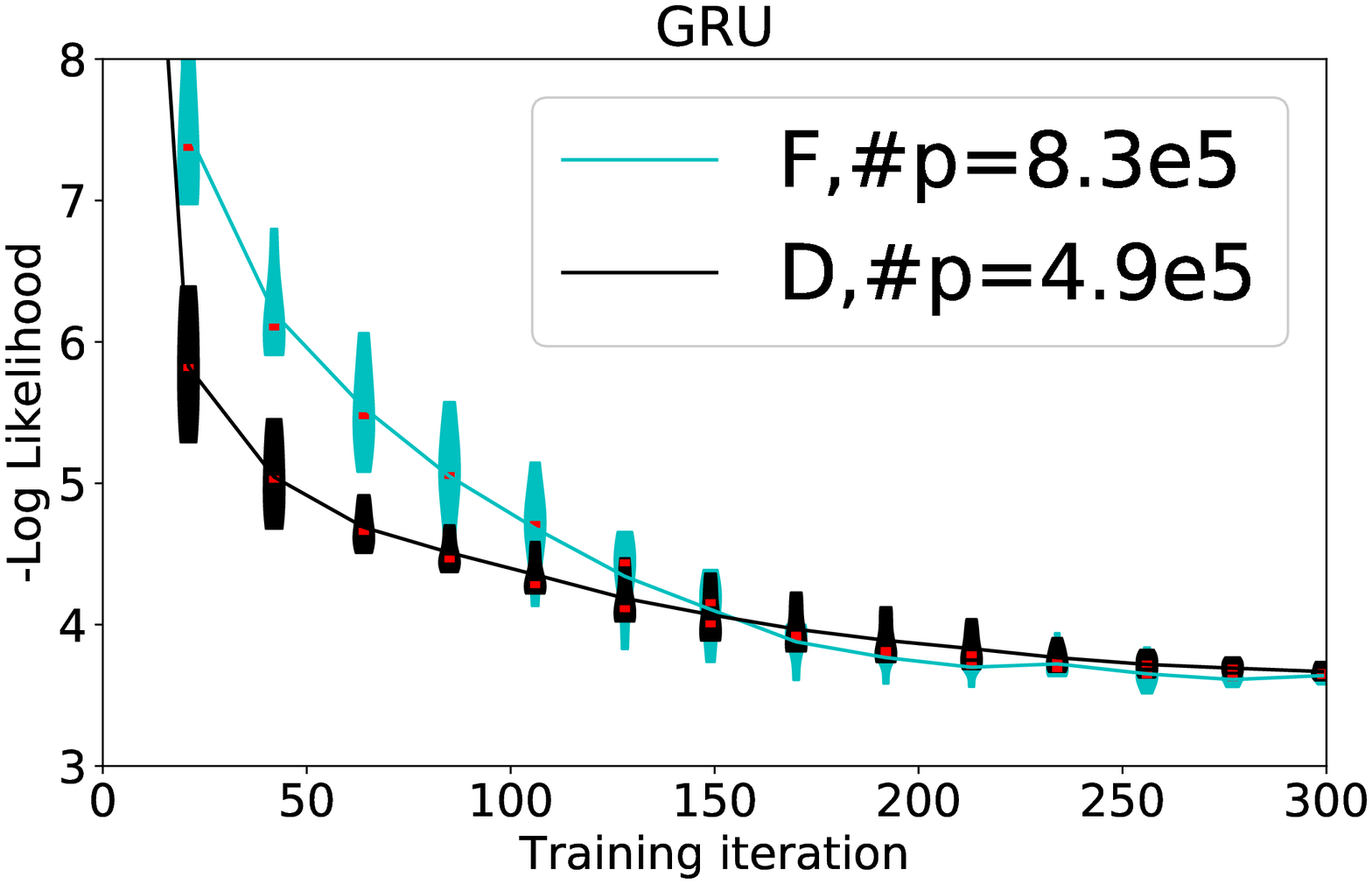} 

  \includegraphics[trim = {\lc cm \bc cm \rc cm \uc cm},clip, width = \sz\textwidth]{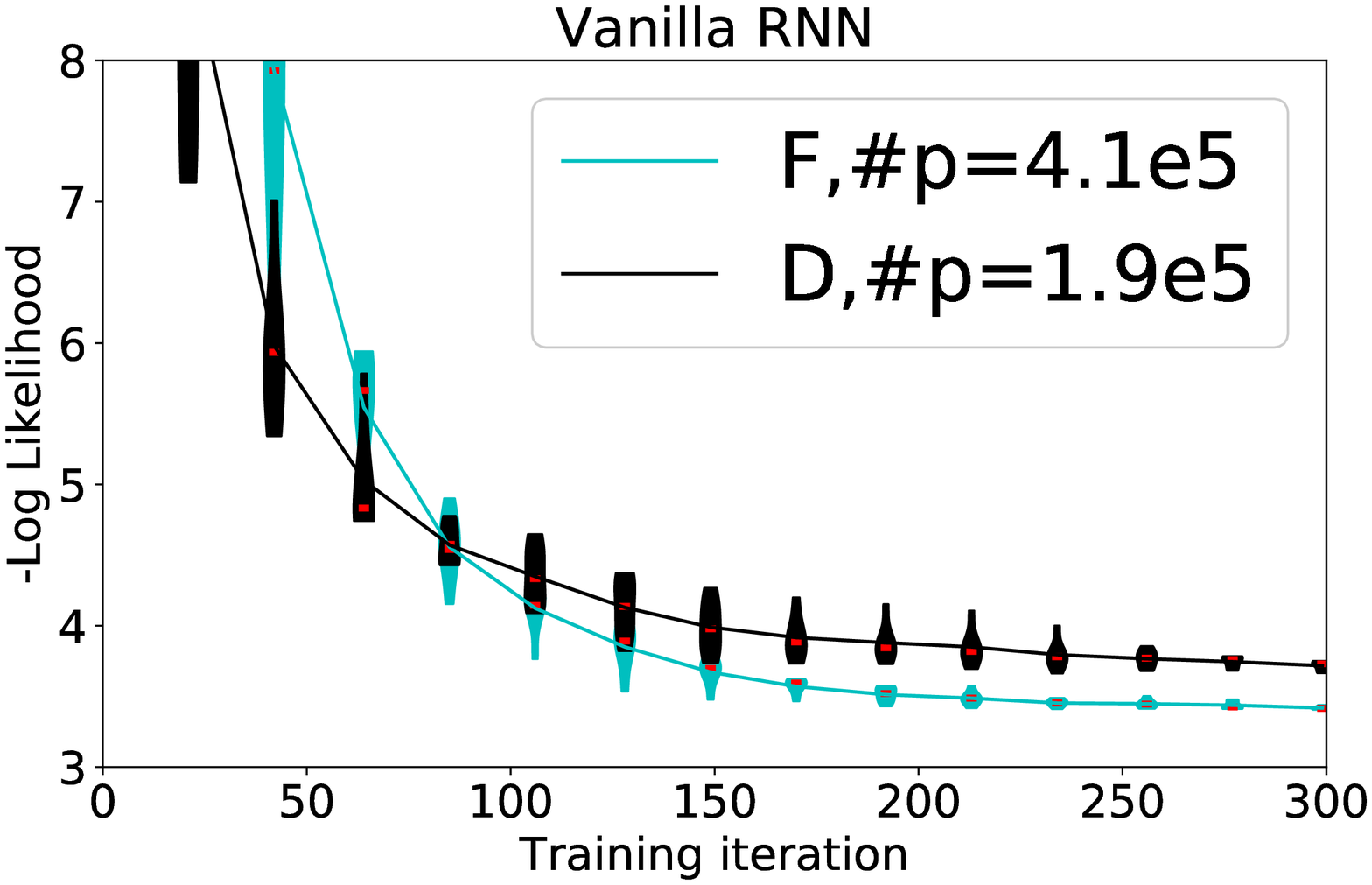} 
  \includegraphics[trim = {\lc cm \bc cm \rc cm \uc cm},clip, width = \sz\textwidth]{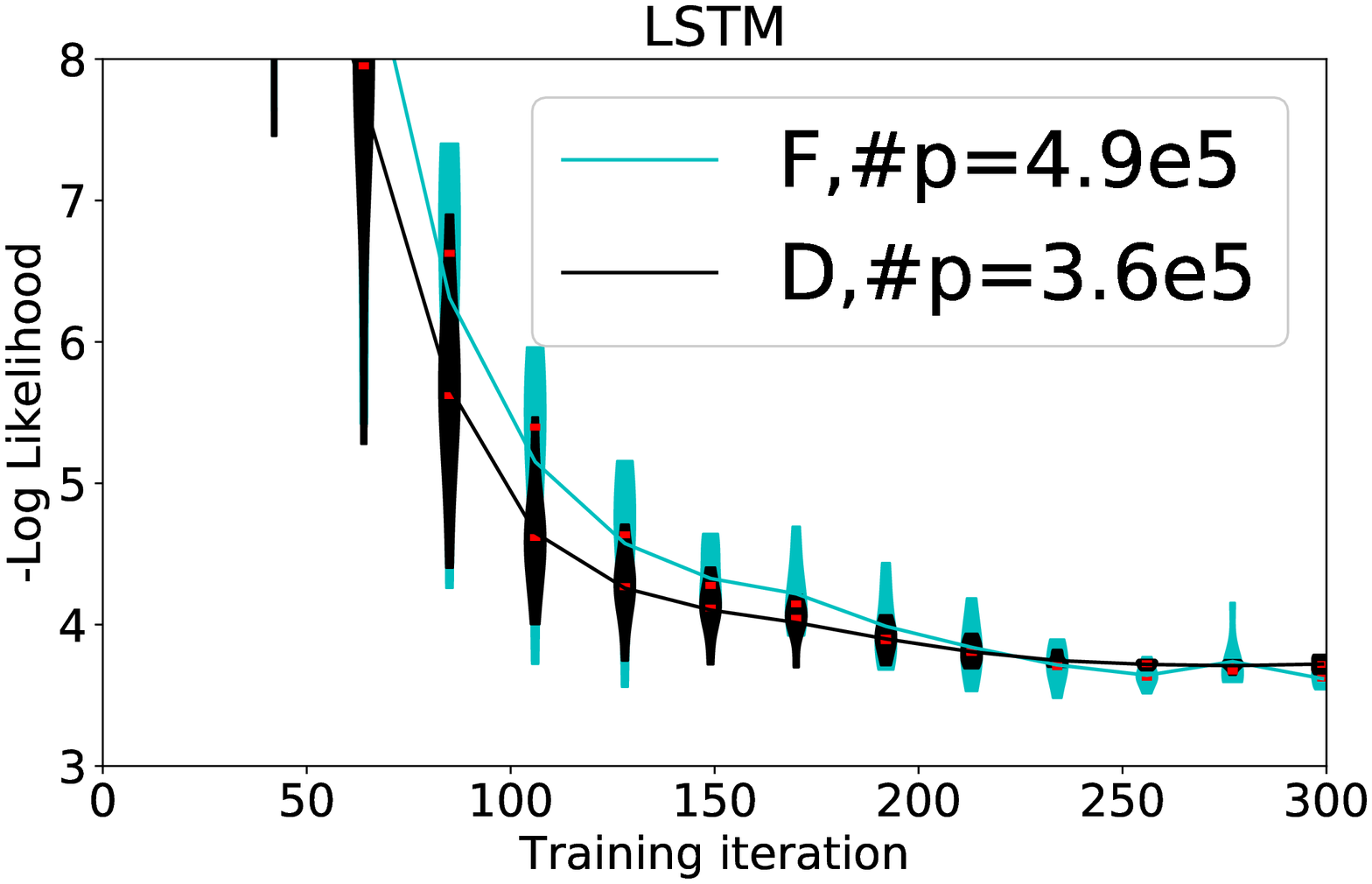} 
  \includegraphics[trim = {\lc cm \bc cm \rc cm \uc cm},clip, width = \sz\textwidth]{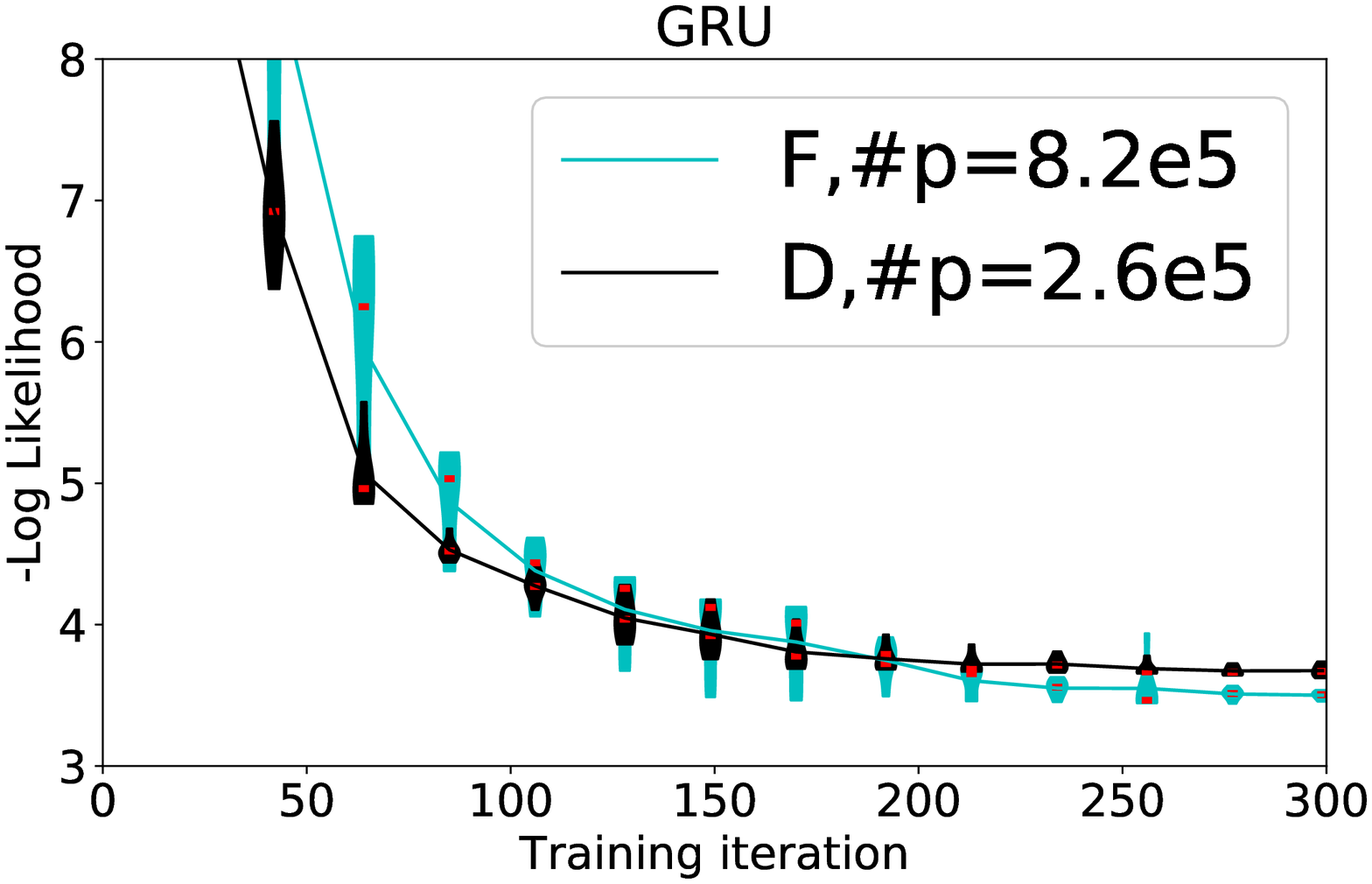} 

  \vspace{-0.2cm}
  \caption{Training iterations vs test negative log-likelihoods on Nottingham dataset.}
  \label{fig:Nottingham}
\end{figure}

\begin{figure}[h!]
  \centering
  \includegraphics[trim = {\lc cm \bc cm \rc cm \uc cm},clip, width = \sz\textwidth]{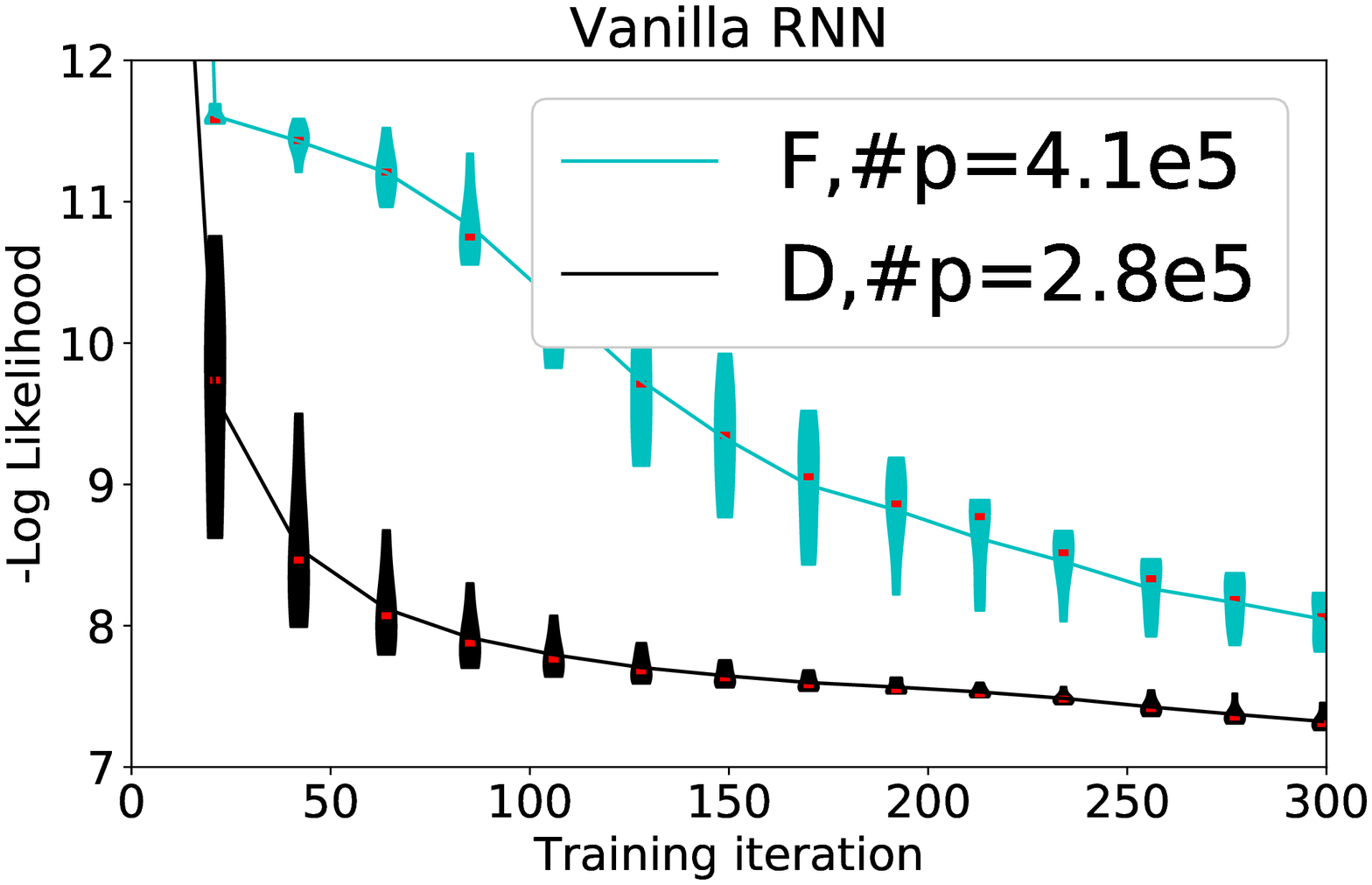}  
  \includegraphics[trim = {\lc cm \bc cm \rc cm \uc cm},clip, width = \sz\textwidth]{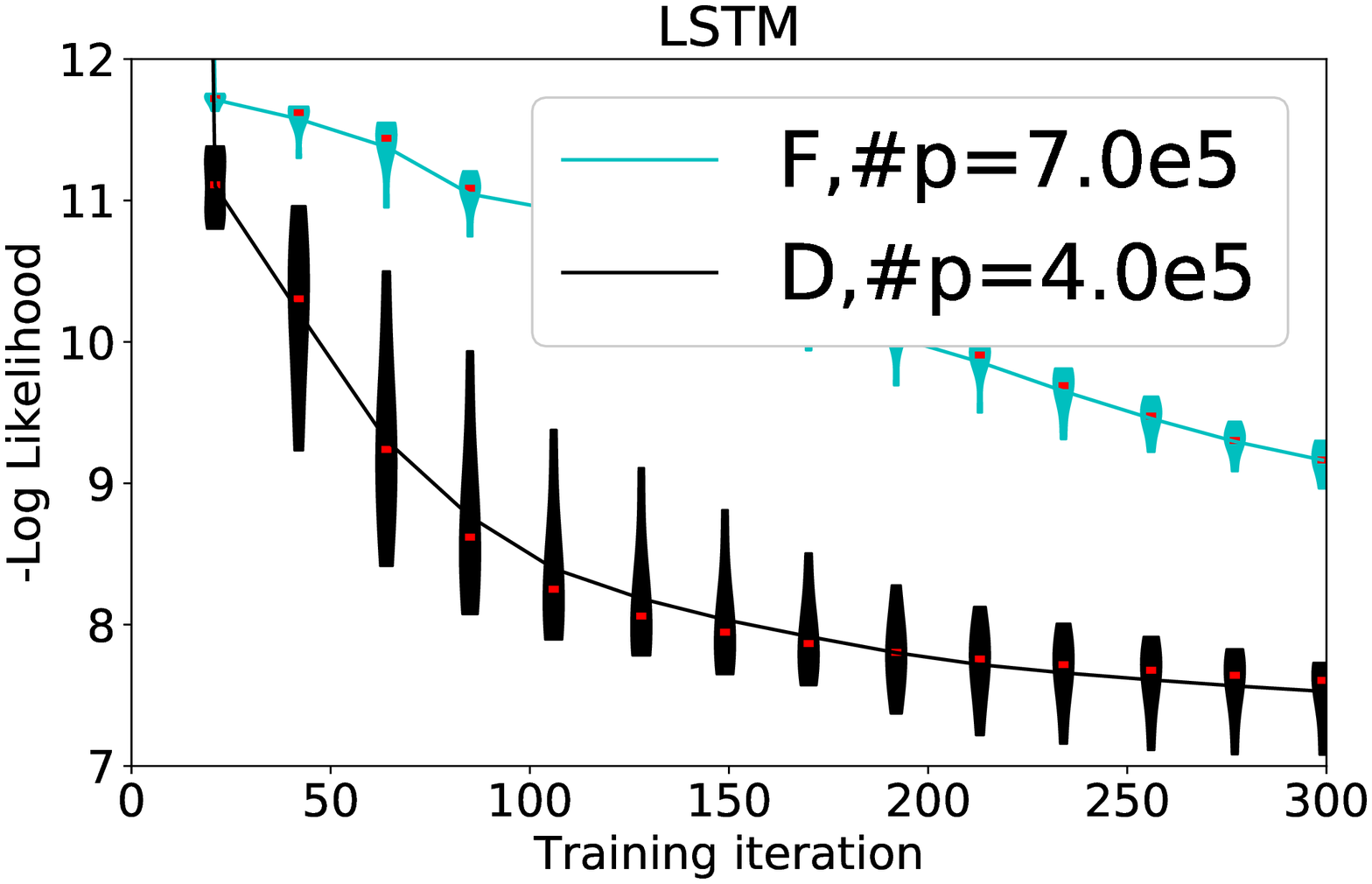} 
  \includegraphics[trim = {\lc cm \bc cm \rc cm \uc cm},clip, width = \sz\textwidth]{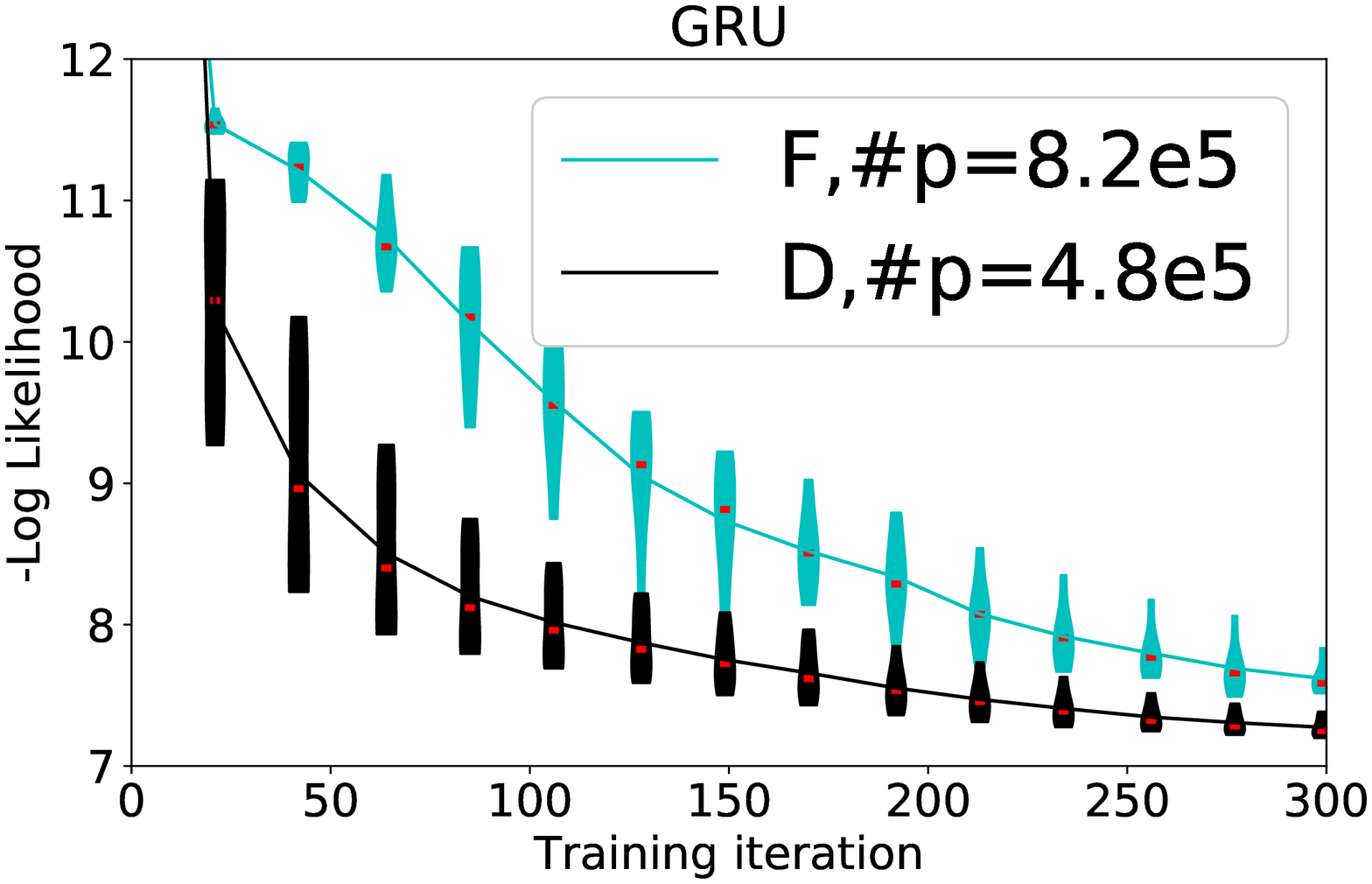} 

  \includegraphics[trim = {\lc cm \bc cm \rc cm \uc cm},clip, width = \sz\textwidth]{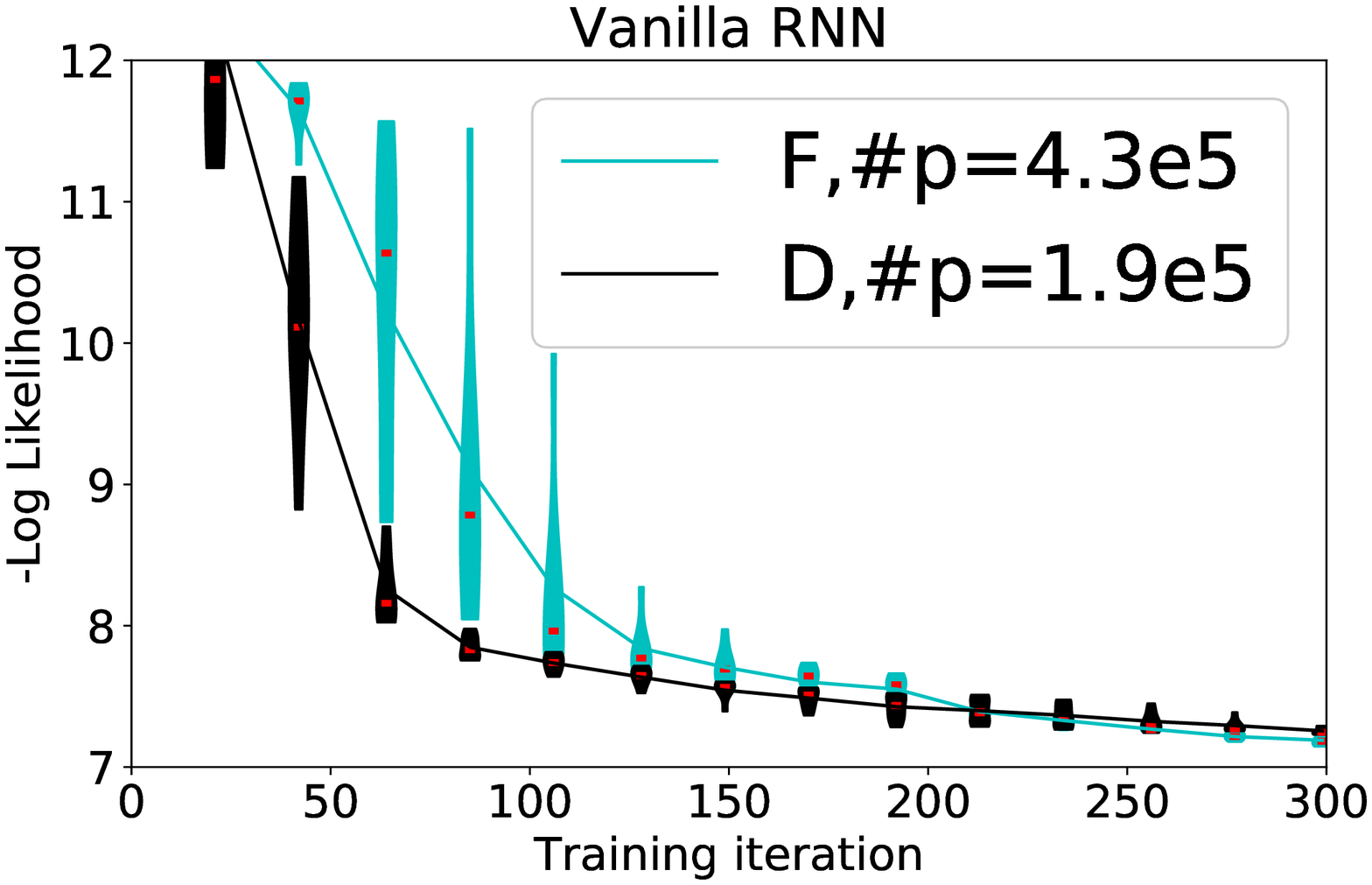} 
  \includegraphics[trim = {\lc cm \bc cm \rc cm \uc cm},clip, width = \sz\textwidth]{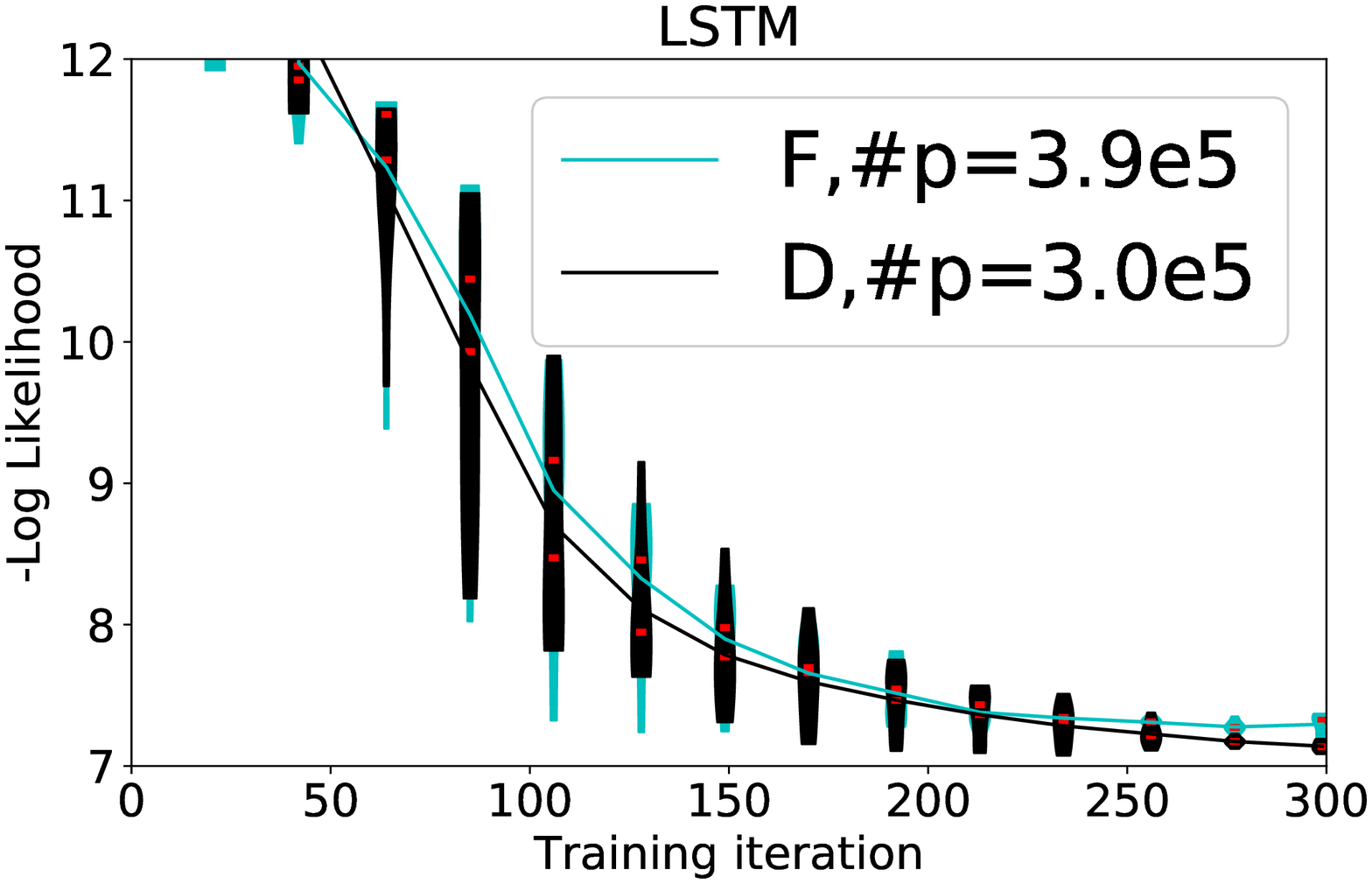} 
  \includegraphics[trim = {\lc cm \bc cm \rc cm \uc cm},clip, width = \sz\textwidth]{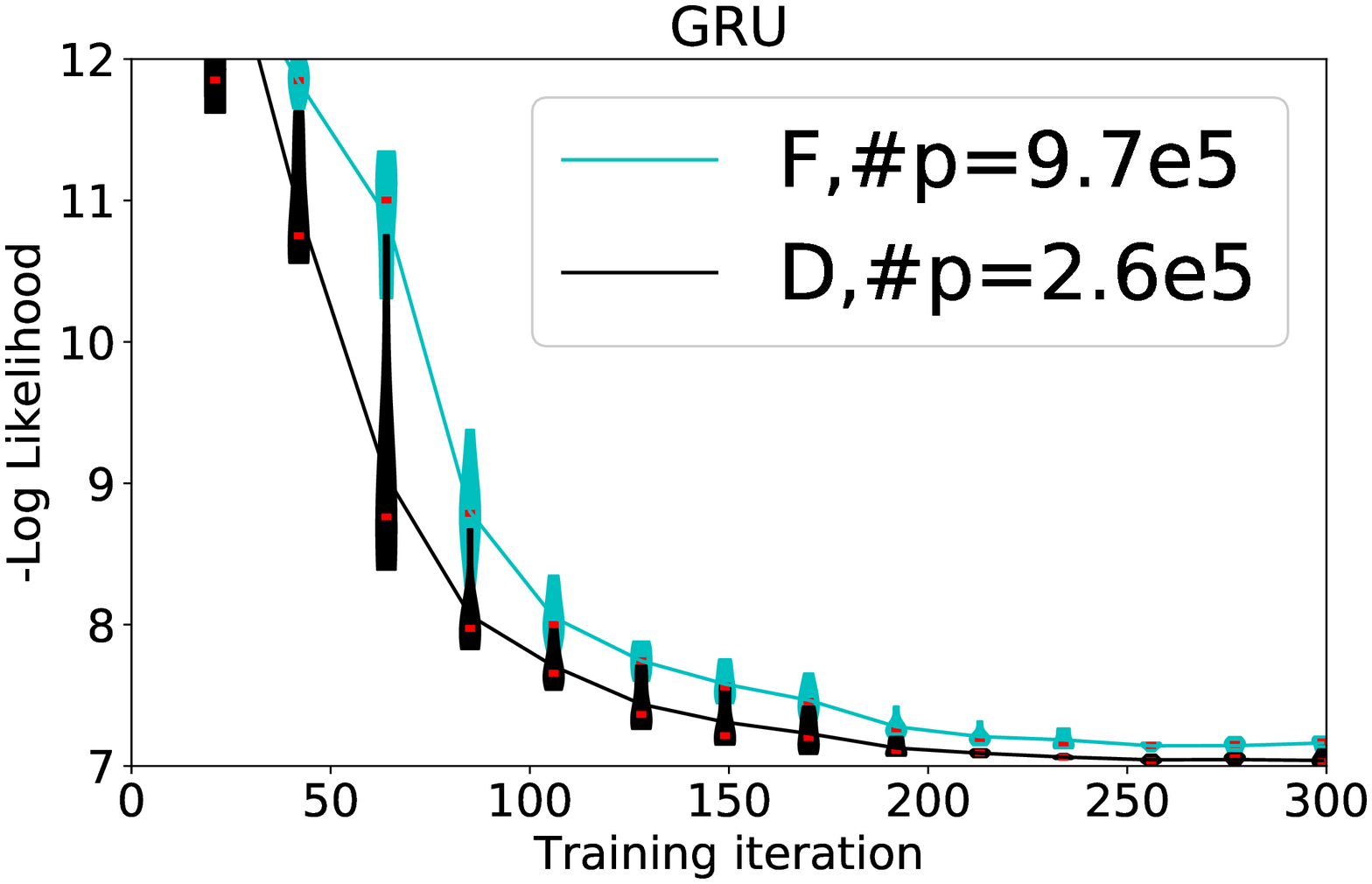} 

  \vspace{-0.2cm}
  \caption{Training iterations vs test negative log-likelihoods on MuseData dataset.}
  \label{fig:MuseData}
\end{figure}
\vspace{-0.34cm}
\section{Conclusions} 
\begin{itemize}
	\item We see that using diagonal recurrent matrices results in an improvement in test likelihoods in almost all cases we have explored in this paper. The benefits are extremely pronounced with the Adam optimizer, but with RMSprop optimizer we also see improvements in training speed and the final test likelihoods. The fact that this modification results in an improvement for three different models and two different optimizers strongly suggests that using diagonal recurrent matrices is suitable for modeling symbolic music datasets, and is potentially useful in other applications.  
	\item Except the Nottingham dataset, using the diagonal recurrent matrix results in an improvement in final test likelihood in all cases. Although the final negative likelihoods on the Nottingham dataset are larger for diagonal models, we still see some improvement in training speed in some cases, as we see that the \diagcolor curves lie below the \fullcolor curves for the most part.   
	\item We see that the average number of parameters utilized by the top 6 diagonal models is in most cases smaller than that of the top 6 full models: In these cases, we observe that the diagonal models achieve comparable (if not better) performance by using fewer parameters. 
\end{itemize}
Overall, in this paper we provide experimental data which strongly suggests that the diagonal RNNs can be a great alternative for regular full-recurrent-matrix RNNs. 

\bibliographystyle{IEEEtran}
\bibliography{refs17}

\end{sloppy}
\end{document}